# A CRITICAL REVIEW OF THE NEED FOR KNOWLEDGE–CENTRIC EVALUATION OF QURANIC RECITATION

A PREPRINT

**Mohammed Hilal Al-Kharusi**[1], **Khizar Hayat***, **Khalil Bader Al Ruqeishi**[1], **and Haroon Rashid Lone**[2]
[1]College of Arts and Sciences, University of Nizwa, Sultanate of Oman.
[2]Ascend Solutions, Riyadh, Kingdom of Saudi Arabia.

November 10, 2025

## ABSTRACT

The art and science of Quranic recitation (*Tajweed*), a discipline governed by meticulous phonetic, rhythmic, and theological principles, confronts substantial educational challenges in today's digital age. Although modern technology offers unparalleled opportunities for learning, existing automated systems for evaluating recitation have struggled to gain broad acceptance or demonstrate educational effectiveness. This literature review examines this crucial disparity, offering a thorough analysis of scholarly research, digital platforms, and commercial tools developed over the past twenty years. Our analysis uncovers a fundamental flaw in current approaches that adapt Automatic Speech Recognition (ASR) systems, which emphasize word identification over qualitative acoustic evaluation. These systems suffer from limitations such as reliance on biased datasets, demographic disparities, and an inability to deliver meaningful feedback for improvement. Challenging these data-centric methodologies, we advocate for a paradigm shift toward a knowledge-based computational framework. By leveraging the unchanging nature of the Quranic text and the well-defined rules of Tajweed, we propose that an effective evaluation system should be built upon rule-based acoustic modeling centered on canonical pronunciation principles and articulation points (*Makhraj*), rather than depending on statistical patterns derived from flawed or biased data. The review concludes that the future of automated Quranic recitation assessment lies in hybrid systems that combine linguistic expertise with advanced audio processing. Such an approach paves the way for developing reliable, fair, and pedagogically effective tools that can authentically assist learners across the globe.

## 1 Introduction

The Qur'an, the foundational sacred book of Islam, holds profound spiritual, legal, and linguistic significance for over two billion Muslims globally. Its recitation (*Tajweed*) transcends mere reading; it is a sacred act governed by precise rules of pronunciation, rhythm, and intonation, as emphasized in the Qur'an: "and recite the Qur'an with measured recitation" (Surah Al-Muzammil, 73:4). The importance of this practice is further underscored in the Prophetic tradition, as the Prophet Muhammad (PBUH) said, "The example of one who recites the Quran and memorizes it is that of one with the 'righteous and noble scribes'. The example of one who recites the Quran and is committed to it, although it is difficult for him, is that of one with a double reward."[Source: Sahih al-Bukhari 4937, Sahih Muslim 798].

Historically, Quranic recitation pedagogy has relied on direct oral transmission from teacher to student (*Mushafa'ha*), a method authenticated through chains of authority (*Esnad*). This rigorous tradition has preserved the Qur'an's authentic pronunciation and meaning for centuries, culminating in formal certification (*Ejazah*) granted only to those demonstrating mastery – a process demanding significant dedication from both student and teacher (*Mujaz*). However, contemporary challenges, including time constraints for learners and teachers, the scarcity of qualified instructors

*khizar.hayat@unizwa.edu.om

(especially outside Muslim-majority regions), and the difficulty of achieving *Ejazah* mastery later in life, create barriers to this traditional path.

The advent of digital technology in the 21st century offers transformative potential. Digital solutions – including mobile applications, websites, online learning platforms, and artificial intelligence (AI) models – now provide unprecedented access to Quranic education, overcoming geographical and temporal limitations. These tools aim to address pressing needs: enhancing accessibility, offering standardized resources for non-native Arabic speakers, and supporting learners striving towards accurate recitation. This integration signifies a critical convergence of faith and modernity, seeking to balance deep reverence for tradition with the practical benefits of digital innovation.

Quranic recitation has persisted as an unbroken oral tradition for over fourteen centuries. While technological advancement has revolutionized educational tools across domains, no computationally robust solution has achieved widespread adoption for automated Tajweed-compliant recitation evaluation. This presents a critical paradox:

Despite the Quran's enduring oral tradition, the development of a computationally robust and widely adopted tool for automated Tajweed-compliant recitation evaluation remains an unresolved paradox. While technological feasibility is demonstrated in other domains, existing Quranic applications fail to achieve widespread adoption due to a critical gap between their availability and their effectiveness. This gap is characterized by insufficient technical rigor in handling Arabic's phonetic complexity and a lack of pedagogical utility needed to gain user trust. Consequently, this review aims to evaluate current applications, synthesize global research from the early 2000s to 2025, and identify critical computational gaps. It will investigate the existence and performance of current systems, analyze the barriers to their adoption, and explore the architectural requirements for developing theologically sound and scalable evaluation tools across the key dimensions of learning tools, preservation efforts, and scholarly research.

Structurally, the review begins by examining the concept of Tajweed and the audio technologies. In the literature section, academic efforts related to the evaluation of Qur'anic recitation will be mentioned. In addition, non-academic efforts, such as websites and applications, will be discussed. The subsequent section critically analyzes contemporary papers, their computational architectures, and user experiences, incorporating evidence from an Arabic point of view. The concluding sections address emergent challenges and outline future research directions, including AI's potential to model expert recitation patterns and develop adaptive learning systems. Crucially, the review assesses how current technological efforts measure against the ideal of supporting accurate mastery.

## 2 Background

This review centers exclusively on the application of technology to Quranic recitation (*Tilawah*) and its associated rules (*Tajweed*), rather than delving into *Tajweed* as a traditional Islamic science. Consequently, this section establishes an essential conceptual foundation rather than offering a comprehensive examination of *Tajweed*.

### 2.1 Tajweed/ Recitation

*Tilawa*, as explained in [1], is the practice of Quranic recitation that inherently follows the rules of *Tajweed*. The term *Tajweed* linguistically means improvement or enhancement. In technical terms, however, it is a Qur'anic discipline that involves pronouncing each letter from its specific point of articulation and ensuring that each letter is given its due phonetic characteristics. Among the many rules of *Tajweed*, some of the most common include:

- **Articulation:** Points of articulation are the specific locations where each letter is pronounced. Arabic scholars categorize these points into four main areas: the open space in the mouth and throat, the throat, the mouth, and the nasal cavity [2]. The mouth articulations are further divided into four specific areas: the hard palate, the tongue, the teeth, and the lips. A visual representation of these articulation points is shown in Figure 1.
- ***Sifat*:** The phonetic attributes of Arabic letters, referred to as *Sifat*, determine how each character is pronounced following its articulation. These attributes are organized into three distinct categories based on their intensity: strong, weak, and intermediate. Additionally, some *Sifat* have opposing characteristics (for example, apparent versus whispering), while others, such as Qalqala, do not possess opposites. Table 1 presents the strong and weak *Sifat* for all letters of the Arabic alphabet.
- ***Waqf* and *Ibtida'*:** The rules of *Waqf* and *Ibtida'* are essential for determining permissible pauses and continuations within Quranic verses. Incorrect pausing during recitation can alter the intended meaning of the text, a concern shared by all languages. In Arabic, this issue is particularly significant due to the use of diacritical marks (*Tashkeel*). Since pauses cannot occur on letters with moving diacritics, *Tajweed* establishes specific guidelines for appropriate stopping and starting points.

| | | Strong charactersitics | | | | | | | | | | | Weak charactersitics | | | |
|---|---|---|---|---|---|---|---|---|---|---|---|---|---|---|---|---|
| Character (Arabic) | Phone (IPA) | Apparent | Strong | Elevated | Closed | Qalqala (vibrated) | Whistled | Drifted | Spreaded | Lengthened | Repeated | Nasalized | Whispered | Soft | Lowered | Opened |
| ء | /ʔ/ | √ | ✓ | | | | | | | | | | | | ✓ | ✓ |
| ب | /b/ | ✓ | ✓ | | | ✓ | | | | | | | | | ✓ | ✓ |
| ت | /t/ | | ✓ | | | | | | | | | | ✓ | | ✓ | ✓ |
| ث | /θ/ | | | | | | | | | | | | ✓ | ✓ | ✓ | ✓ |
| ج | /d͡ʒ/[f] | ✓ | ✓ | | | ✓ | | | | | | | | | ✓ | ✓ |
| ح | /ħ/ | | | | | | | | | | | | ✓ | ✓ | ✓ | ✓ |
| خ | /x/ | | | ✓ | | | | | | | | | ✓ | ✓ | | ✓ |
| د | /d/ | ✓ | ✓ | | | ✓ | | | | | | | | | ✓ | ✓ |
| ذ | /ð/ | ✓ | | | | | | | | | | | | ✓ | ✓ | ✓ |
| ر | /r/ | ✓ | | | | | | ✓ | | | ✓ | | | | ✓ | ✓ |
| ز | /z/ | ✓ | | | | | ✓ | | | | | | | ✓ | ✓ | ✓ |
| س | /s/ | | | | | | ✓ | | ✓ | | | | ✓ | ✓ | ✓ | ✓ |
| ش | /ʃ/ | | | | | | | | | | | | ✓ | ✓ | ✓ | ✓ |
| ص | /sˤ/ | | | ✓ | ✓ | | ✓ | | | | | | ✓ | ✓ | | |
| ض | /dˤ/ | ✓ | | ✓ | ✓ | | | | | ✓ | | | | ✓ | | |
| ط | /tˤ/ | ✓ | ✓ | ✓ | ✓ | ✓ | | | | | | | | | | |
| ظ | /ðˤ/ | ✓ | | ✓ | ✓ | | | | | | | | | ✓ | | |
| ع | /ʕ/ | ✓ | | | | | | | | | | | | | ✓ | ✓ |
| غ | /ɣ/ | ✓ | | ✓ | | | | | | | | | | ✓ | | ✓ |
| ف | /f/ | | | | | | | | | | | | ✓ | ✓ | ✓ | ✓ |
| ق | /q/ | ✓ | ✓ | ✓ | | ✓ | | | | | | | | | | ✓ |
| ك | /k/ | | ✓ | | | | | | | | | | ✓ | | ✓ | ✓ |
| ل | /l/ | ✓ | | | | | | ✓ | | | | | | | ✓ | ✓ |
| م | /m/ | ✓ | | | | | | | | | | ✓ | | | ✓ | ✓ |
| ن | /n/ | ✓ | | | | | | | | | | ✓ | | | ✓ | ✓ |
| ه | /h/ | | | | | | | | | | | | ✓ | ✓ | ✓ | ✓ |
| و | /w/ | ✓ | | | | | | | | | | | | ✓ | ✓ | ✓ |
| ا | /aː/[e] | ✓ | | | | | | | | | | | | ✓ | ✓ | ✓ |
| ي | /j/ | ✓ | | | | | | | | | | | | ✓ | ✓ | ✓ |

Table 1: *Sifat* according to [2]

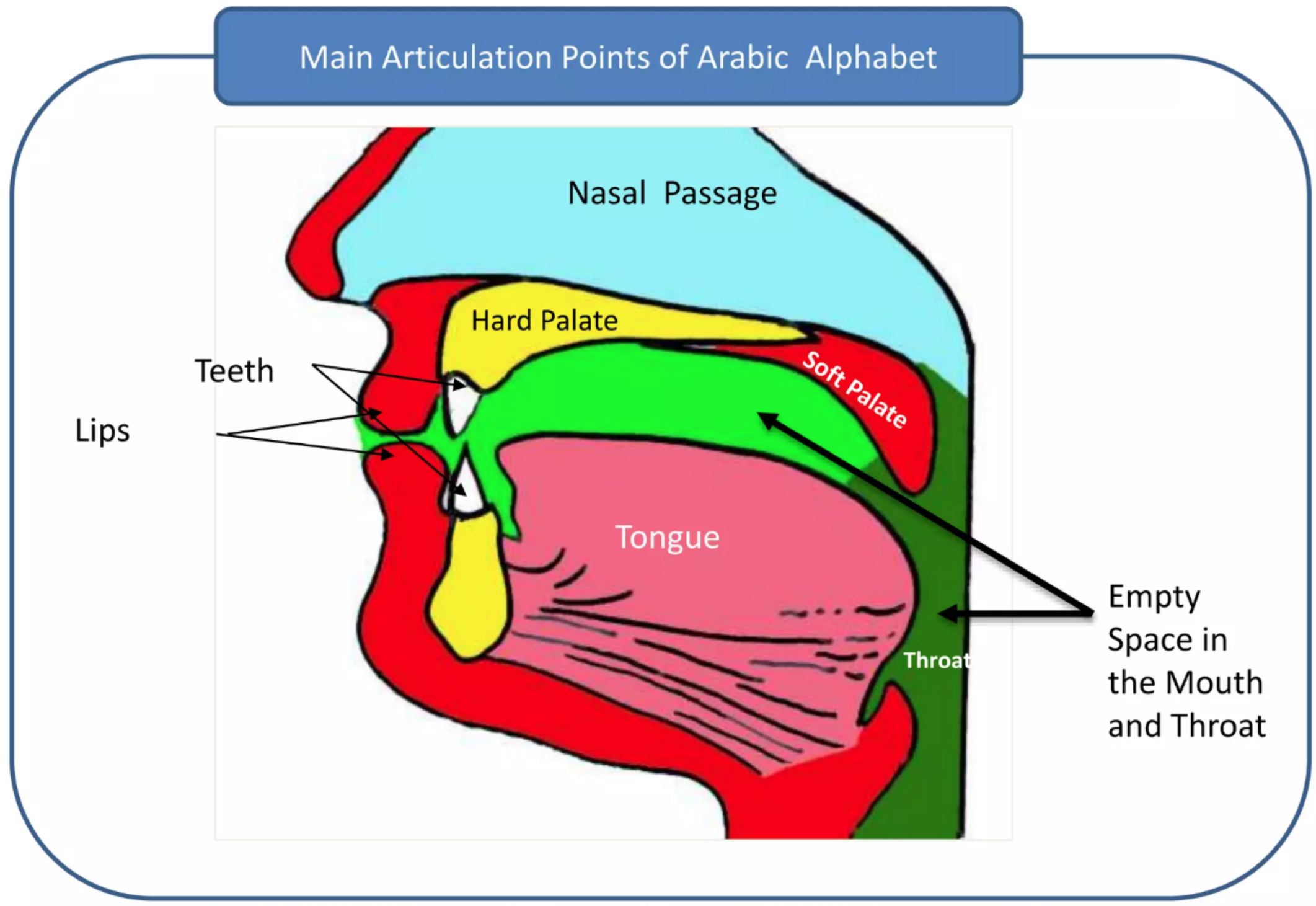

Figure 1: Makharj summary [3]

- ***Meem and Noon Rules*:** The letters Meem (م) and Noon (ن) possess unique phonetic properties due to their nasal characteristics. Consequently, special rules apply when these letters appear with a static diacritical mark (*Sakin*).

- ***Raa Rules*:** The pronunciation of the letter Raa (ر) can vary in weight, depending on its diacritical marking. Therefore, *Tajweed* provides specific criteria to determine whether Raa should be pronounced heavily or lightly within a given context.

- ***Elongation (Maad)*:** Elongation in Arabic, referred to as *Maad*, encompasses various types influenced by multiple factors. These include the diacritical mark of the subsequent character and its position within the sentence. Elongation occurs when a vowel diacritic is followed by a letter with a static diacritical mark (*Sokoon*), such as *Fatha* with ا, *Dhamma* with و, and *Kasra* with ي.

- ***Duration (Harakah)*:** In *Tajweed*, the concept of *Harakah* pertains to the duration for which each letter or rule should be sustained. For instance, certain types of *Maad* require a longer duration compared to *Ghunnah*, which typically lasts for two *Harakah*.

Understanding the transmission of *Tajweed* necessitates familiarity with the method of succession, known as *Tawatur*. A canonical method of Quranic recitation, referred to as *Qira'ah*, is defined as the recitation attributed to an Imam from one of the ten recognized reciters (*Qurra'*, the plural of *Qari*) [4]. These reciters received their recitation directly from the Prophet Muhammad (peace be upon him), thereby preserving his original method of recitation. A narration, or

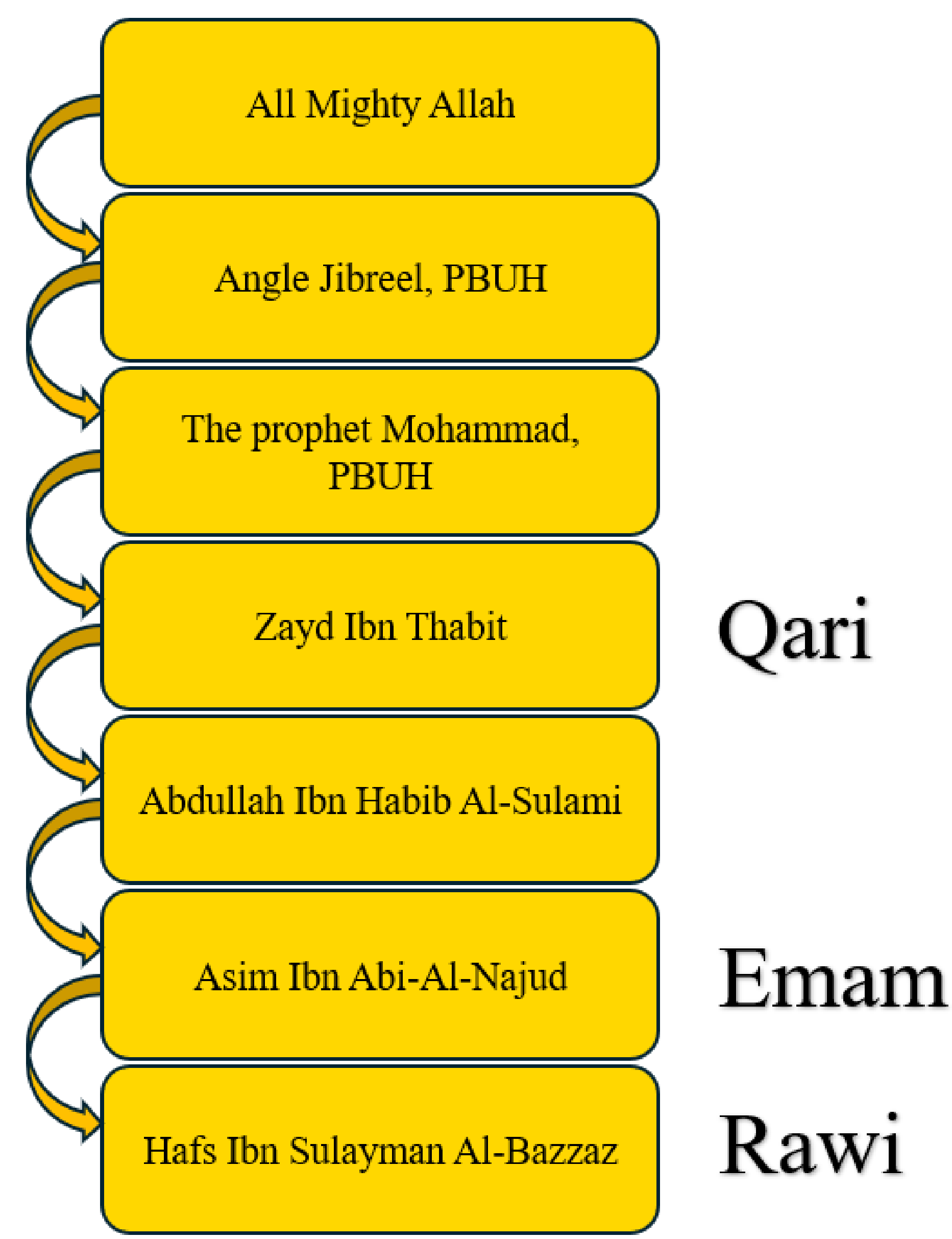

Figure 2: Rewaya of Hafs form Asim

*Riwayah*, pertains to the specific transmission attributed to a narrator (*Rawi*) from the Imam [4]. The chain of narration for *Hafs* from *Asim* is illustrated in Figure 2.

Mastering Quranic recitation is formally recognized through a certification known as *Ijaza*. A certified teacher, referred to as a *Mujiz*, holds this *Ijaza* and is thereby authorized to instruct and confer certification upon others. A critical aspect of the *Ijaza* is its documentation of an unbroken chain of transmission, or *Sanad*, tracing back to the Prophet Muhammad (peace be upon him). This unbroken chain is fundamental to preserving the authenticity of the recitation. This systematic approach ensures the protection of recitation from any alterations, as *Ijazas* are exclusively granted by authorized individuals after thorough verification. Any interruption or weakness in the *Sanad* renders the authorization invalid and disqualifies the recipient from granting further *Ijazas* [4].

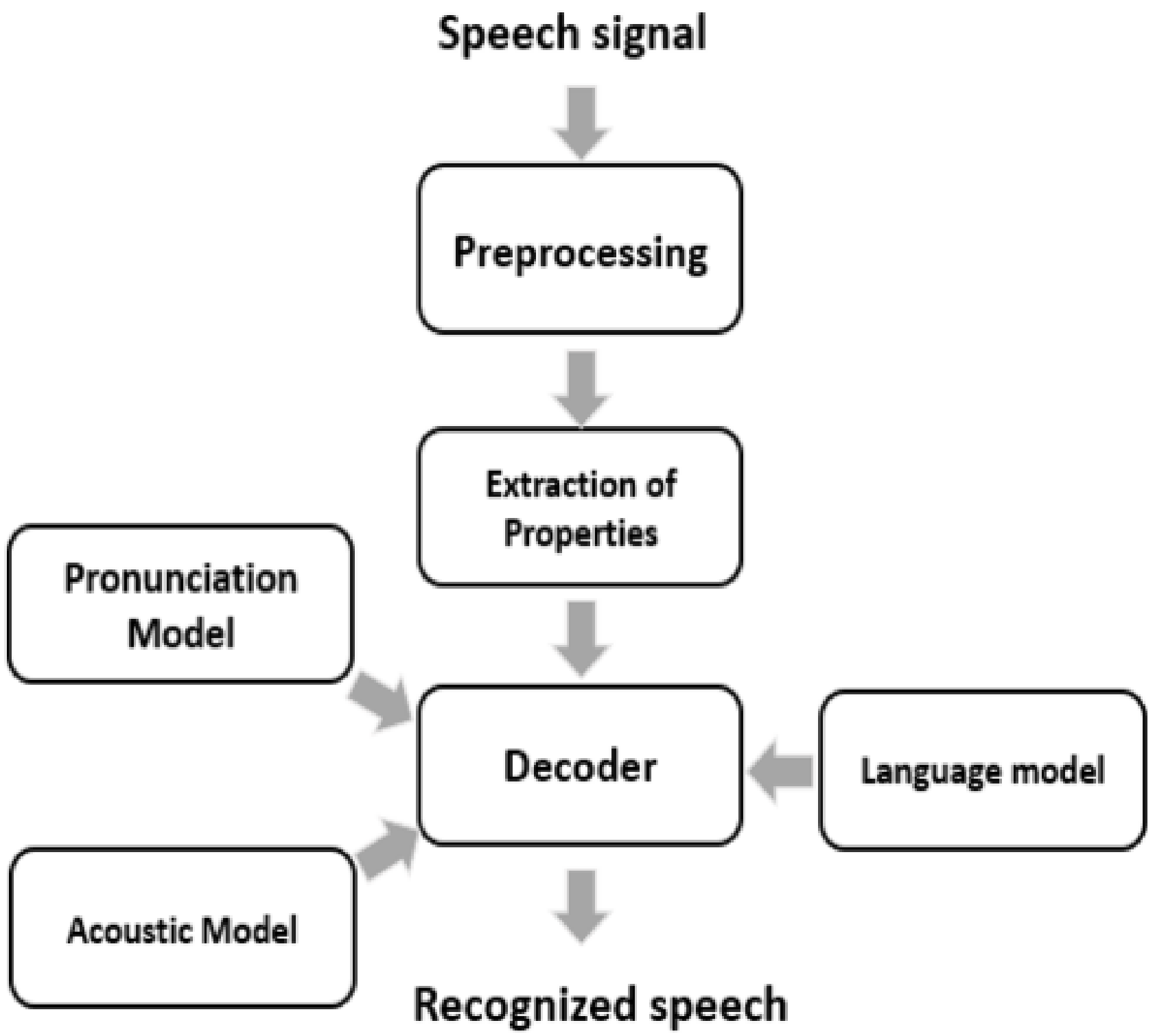

Figure 3: ASR model [5]

### 2.2 Automatic Speech Recognition

Automatic Speech Recognition (ASR) offers a computational approach to converting spoken language into written text. This approach comprises several interconnected components (as illustrated in Figure 3), which work in unison to achieve accurate recognition. Key stages—including signal preprocessing, feature extraction, acoustic modeling, language modeling, and decoding—will be explored in detail below. Figure 3 provides an overview of the overall structure of ASR systems.

#### 2.2.1 Digitalizing Audio:

Sound travels as a continuous wave through a medium and becomes audible to humans when its frequency lies within the audible range, typically between 20 Hz and 20,000 Hz. For computational analysis, this analog signal must be converted into a discrete form by sampling its amplitude at regular intervals. The sampled signal is then digitized into a numerical format, making it suitable for processing and storage in digital systems. This conversion is generally carried out using microphones or other audio capture devices. Once in digital form, the audio data can be processed computationally [6].

#### 2.2.2 Preprocessing:

Audio preprocessing enhances the quality of the signal through various techniques, such as noise reduction and silence removal. Commonly used algorithms for this purpose include Spectral Subtraction and Minimum Mean-Square Error Short-Time Spectral Amplitude (MMSE-STSA) estimation for noise suppression, as well as Short-Time Energy (STE) for detecting and eliminating silent segments [7].

#### 2.2.3 Segmentation:

After preprocessing, audio signals are divided into smaller segments to isolate relevant components for further analysis. The level of segmentation depends on the specific application, with common approaches including the division of utterances into sentences, words, or sub-word units. Most frequently, segmentation is performed at the phoneme level, as phonemes represent the smallest units of speech sounds modeled in acoustic frameworks. These segmented samples are then used as input for the subsequent feature extraction stage [8].

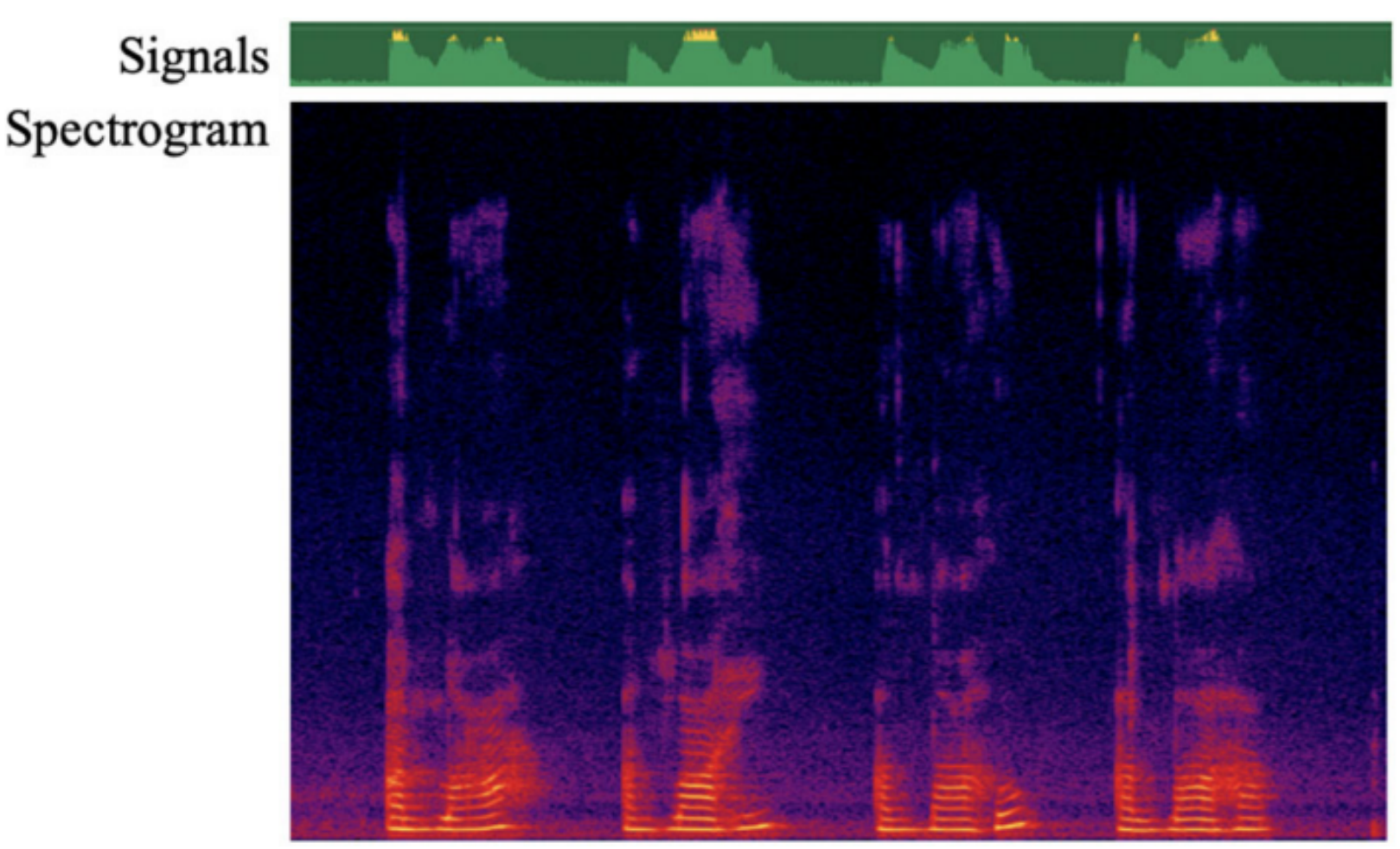

Figure 4: Spectrogram [9]

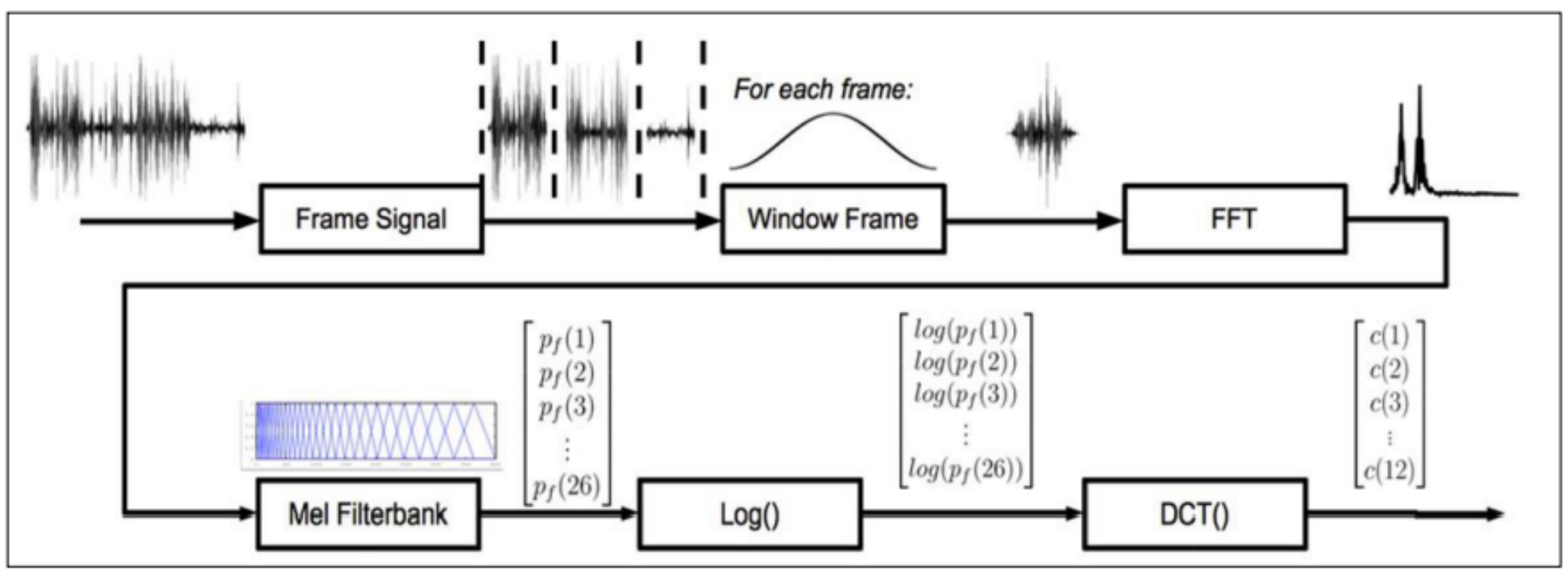

Figure 5: MFCC steps [10]

#### 2.2.4 Feature Extraction:

Feature extraction involves converting segmented audio samples into meaningful representations that can be effectively modeled. One of the most common initial representations is the spectrogram, which is generated using the Short-Time Fourier Transform (STFT). A spectrogram visually depicts the energy of the signal across different frequency bands over time, as illustrated in Figure 4 [6]. Mel Frequency Cepstral Coefficients (MFCCs) are the most widely used features in acoustic modeling, particularly in the field of Automatic Speech Recognition (ASR) [11]. The process of deriving MFCCs involves multiple stages, which are summarized in Figure 5:

1. Framing and applying a window function to the signal.
2. Calculating the magnitude of the Short-Time Fourier Transform (STFT).
3. Applying Mel-scale filterbanks to emulate human auditory perception.
4. Taking the logarithm of the filterbank energies.
5. Applying the Discrete Cosine Transform (DCT) to decorrelate the filterbank outputs, resulting in cepstral coefficients.

In addition to MFCCs, there are numerous alternative feature extraction techniques, each with unique characteristics and suitability for different model architectures. Table 2 presents some of the most commonly used features in acoustic modeling.

| Feature Type | Description | Primary Usage |
|---|---|---|
| MFCC | Coefficients representing spectral envelope shape via Mel-scaled filterbanks and Discrete Cosine Transform (DCT). | Dominant baseline for GMM-HMM and hybrid DNN-HMM Automatic Speech Recognition (ASR) systems. |
| Filterbank Energies (FBANK) | Log Mel-scaled filterbank energies (includes MFCC steps 1-4 but omits DCT transformation). | Preferred input representation for modern Deep Neural Network (DNN) and Recurrent Neural Network (RNN) acoustic models. |
| Perceptual Linear Prediction (PLP) | Applies psychophysical concepts including Bark scale and loudness equalization combined with linear prediction analysis. | Provides robustness in noisy environments; commonly used in traditional HMM-based speech recognition systems. |
| Delta & Delta-Delta MFCCs | First and second temporal derivatives of MFCCs, designed to capture dynamic speech features and temporal transitions. | Augments static MFCC features to improve contextual modeling in speech recognition tasks. |
| Linear Predictive Coding (LPC) | Models speech production mechanism using linear prediction coefficients (LPCs) or derived cepstral coefficients (LPCC). | Primarily used in older speech coding and recognition systems with focus on speech synthesis applications. |
| Gammatone Frequency Cepstral Coefficients (GFCC) | Utilizes gammatone filterbanks that mimic cochlear response patterns instead of traditional Mel scaling. | Offers improved robustness in noisy conditions; used in auditory-motivated ASR research. |
| Power-Normalized Cepstral Coefficients (PNCC) | Applies auditory-inspired power-law nonlinearity combined with medium-time processing techniques. | Designed for high noise robustness; particularly effective in challenging acoustic environments. |
| Spectral Features (e.g., Spectral Centroid, Flux) | Direct measurements of spectral distribution characteristics and temporal changes in spectral content. | Often used as supplementary features for emotion recognition and paralinguistic analysis tasks. |
| Learned Features from raw data (e.g., Wav2Vec 2.0, HuBERT) | Deep neural networks trained to extract latent representations directly from raw or minimally processed audio signals. | Represents state-of-the-art approach for end-to-end and transformer-based ASR systems. |

Table 2: Comparison of feature extraction methods commonly used in speech processing and automatic speech recognition systems

#### 2.2.5 Acoustic Model:

In the ASR pipeline, acoustic models serve as classifiers that process extracted audio features, such as MFCCs or filterbank energies. Their core function is to map sequences of these features to their most likely corresponding linguistic units, which are typically phonemes in the context of Tajweed-based ASR systems. By analyzing spectral and temporal patterns, these models learn to recognize the unique acoustic signatures associated with each phoneme or sound unit, such as specific *Makhraj* or certain *Tajweed* rules. The output of this stage consists of a sequence of probability distributions over linguistic units for each input feature frame. These probabilities, which reflect the model's confidence in identifying specific sounds, are subsequently used in the decoding phase. During decoding, these probabilities are combined with language models to produce the final text transcription. Language models, which capture word sequences and grammatical structures, will be discussed in the following section.

One of the foundational acoustic models is the Hidden Markov Model (HMM), which provides the temporal framework for modeling speech in traditional ASR systems. An HMM represents speech as a sequence of hidden states, such as phonemes or sub-phonetic units, where each state probabilistically generates observable acoustic features [12]. HMMs incorporate two key components:

1. **State Transitions:** The probability of transitioning between states, capturing the temporal progression of speech.
2. **Emission Probabilities:** The likelihood of observing specific acoustic features (e.g., MFCC vectors) given a particular state.

To model the complex spectral variability within each state—such as variations in pronunciation of a letter across different speakers—Gaussian Mixture Models (GMMs) are utilized. A GMM represents the emission probability distribution of a state as a weighted sum of multiple Gaussian distributions. Each Gaussian component corresponds to a distinct acoustic pattern, such as a specific spectral envelope shape for *Ghunnah* during *Nun Sakinah*.

The combined GMM-HMM framework operates as follows:

1. **Input:** A sequence of acoustic feature vectors (e.g., 39-dimensional MFCCs, including delta and delta-delta features).
2. **HMM State Sequence:** Models the temporal progression of speech units, such as *Makharij*.
3. **GMM Emission Probability:** Computes the probability of observing the input features given the current state.
4. **Output:** Probability scores for hypothesized state sequences, which are then processed by the Viterbi algorithm to determine the most likely path (refer to Figure **??** for an illustration).

Additional acoustic models are summarized in Table 3.

| Acoustic Model | Key Mechanism | Primary Usage |
|---|---|---|
| GMM-HMM | Combines Gaussian Mixture Models (GMMs) to model spectral features per state with Hidden Markov Models (HMMs) for temporal transitions. | Traditional ASR systems; low-resource environments; applications requiring explicit duration modeling (e.g., *Tajweed* elongation rules). |
| DNN-HMM (Hybrid) | Deep Neural Networks (DNNs/CNNs/RNNs) replace GMMs to estimate HMM state probabilities using high-dimensional features. | Modern ASR benchmarks; systems requiring robustness to speaker/noise variations; foundational for *Tajweed* phoneme discrimination. |
| HMM-CTC | Connectionist Temporal Classification (CTC) layer integrated with HMMs to enable sequence prediction without frame-level alignments. | Early end-to-end systems; tasks with variable input-output alignments (e.g., continuous Quranic recitation). |
| RNN Transducer (RNN-T) | Jointly trains acoustic encoder (RNN/CNN), prediction network (RNN for label history), and joint network for symbol prediction. | Streaming ASR (e.g., live *Tajweed* feedback); low-latency applications; state-of-the-art end-to-end systems. |
| Transformer Acoustic Models | Self-attention mechanisms model global dependencies in features; often combined with CTC or RNN-T losses. | Large-scale ASR; complex linguistic contexts (e.g., cross-word *Tajweed* rules like *Idgham*); multilingual systems. |
| TDNN (Time-Delay NN) | 1D convolutional layers capture fixed temporal contexts at different time scales (e.g., +/-5 frames). | Phone recognition in Kaldi pipelines; efficient hardware deployment; dialect-specific *Tajweed* analyzers. |
| Conformer | Hybrid CNN-Transformer architecture: CNN for local features + self-attention for global context. | High-accuracy transcription; long-form recitation analysis; Quranic ASR with nuanced *Makharij* detection. |

Table 3: Comparison of acoustic models used in speech recognition systems

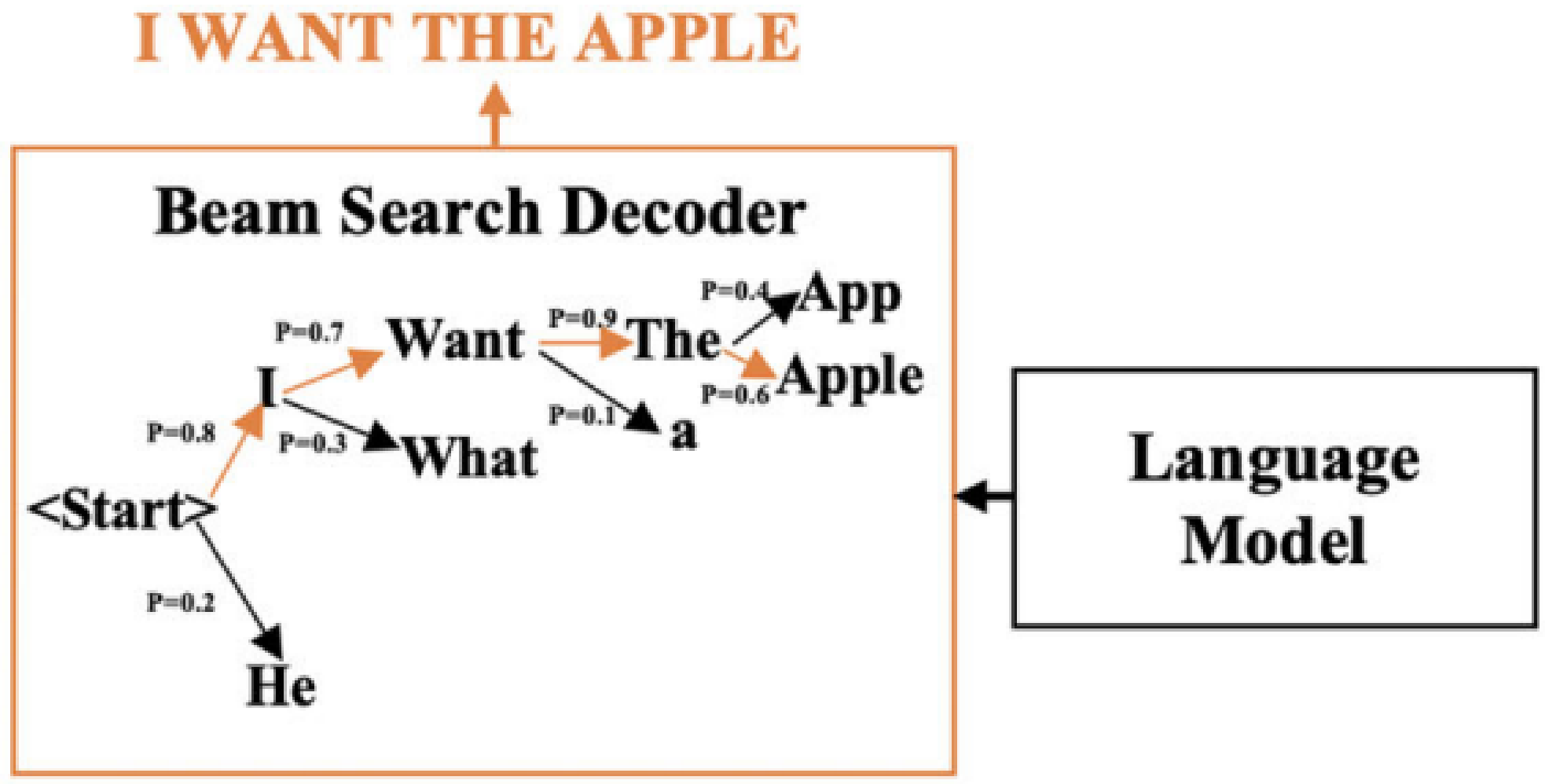

Figure 6: Language model scoring [9]

### 2.2.6 Language Model:

Language models (LMs) play a pivotal role within the ASR framework by providing probability distributions over sequences of words. Their core function is to encode linguistic knowledge—including grammatical structures, common word patterns, and domain-specific semantics—to resolve ambiguities that arise from acoustic similarities. For example, an LM would assign a higher probability to the Quranic phrase بسم الله الرحمن الرحيم (Bismillah al-Rahman al-Rahim) compared to acoustically similar but contextually implausible sequences [12]. This capability is essential for:

1. **Disambiguation:** Differentiating between homophones (e.g., قَلْب [heart] vs. كَلْب [dog]), where acoustic models alone may be insufficient.
2. **Contextual Prediction:** Anticipating likely word sequences based on preceding context (e.g., after إِيَّاكَ نَعْبُدُ, the phrase وَإِيَّاكَ نَسْتَعِينُ in Surah Al-Fatihah would have a high probability).

Traditional n-gram LMs calculate probabilities based on word co-occurrence statistics derived from training corpora [13]. In contrast, modern neural LMs—such as those based on Recurrent Neural Networks (RNNs) or Transformers—leverage distributed representations to model long-range dependencies and rare phrases more effectively [14, 15]. In Tajweed-focused ASR systems, LMs are typically trained on Quranic text corpora that incorporate variations in Qira'at and verse-specific constraints. This ensures that predictions align with canonical recitation rules. LM probabilities are computed to determine the optimal next word, as illustrated in Figure 6. This enables accurate recognition of complex recitation phenomena based on linguistic context alone. Some language models can be seen in the table (table 4) below:

### 2.2.7 Pronunciation Model:

The pronunciation model, typically implemented as a pronunciation lexicon or dictionary, acts as a crucial intermediary between the acoustic model and the language model within the ASR system. Its primary role is to establish a mapping between words (the output of the language model) and their corresponding sequences of sub-word phonetic units (the output of the acoustic model). This component is essential for converting the sequence of recognized phonemes into a coherent sequence of words. Unlike the decoder, which operates dynamically, the pronunciation model functions statically, akin to a dictionary [16].

### 2.2.8 Decoder:

The decoder functions as the search mechanism that integrates probabilities derived from the acoustic model (audio) and the language model (context) to produce the optimal text transcription. Its core task is to navigate the hypothesis space of possible word sequences, balancing two key factors:

| Language Model | Key Mechanism | Key Advantages |
|---|---|---|
| n-gram | Estimates word probability using fixed-length histories (e.g., trigrams). | • Efficient for Qur'an's finite vocabulary<br>• Easily trained on verse databases<br>• Fast prediction for real-time applications |
| RNN-LM | Recurrent neural networks model context via hidden states with memory. | • Captures long verse dependencies (e.g., across pauses/prostration markers)<br>• Handles repetitive liturgical phrases |
| Transformer LM | Self-attention weights importance across all words in the sequence. | • Models complex cross-word *Tajweed* rules (e.g., letter assimilation across words, nasalization effects)<br>• Handles rare phrase combinations |
| Constrained Grammar LM | Enforces pre-defined syntactic/semantic rules. | • Guarantees output matches Quranic text structure<br>• Prevents non-canonical word sequences<br>• Enforces verse boundaries |
| Pronunciation-Augmented LM | Incorporates phonetic variants (e.g., reciter differences). | • Supports legitimate recitation variations (e.g., differing vowel lengths, consonant pronunciations)<br>• Adapts to narrator-specific rules |

Table 4: Comparison of language models used in speech recognition systems

1. **Acoustic Model:** The confidence in recognizing speech units (e.g., phonemes such as ق (Qaf) with correct *Makhraj*).
2. **Language Model:** The likelihood of linguistic sequences (e.g., Arabic sentence structure).

This balance is mathematically represented through the fundamental ASR equation of conditional probability [16].

Key decoding algorithms include:

1. **Viterbi Decoding:** Identifies the single best state sequence in HMM-based systems, which is crucial for precise alignment of *Tajweed* timing (e.g., verifying the duration of *Madd* elongation).
2. **Beam Search:** Efficiently eliminates low-probability hypotheses while maintaining multiple candidates, which is essential for handling variations in *Qira'at* or ambiguous *Idgham* assimilations.
3. **End-to-End Decoders:** Utilize neural sequence-to-sequence models (e.g., attention-based beam search) to directly map features to words, thereby reducing the complexity of the processing pipeline.

In the context of Quranic audio, decoders resolve ambiguities by jointly optimizing acoustic and linguistic evidence, enabling accurate transcription of semantically coherent, *Tajweed*-compliant recitations.

Unlike traditional ASR systems, deep acoustic models leverage neural architectures (e.g., DNNs, CNNs, RNNs, Transformers) to directly learn discriminative representations from raw or minimally processed audio features (e.g., FBANK, spectrograms), thereby replacing the decoder block with a deep acoustic model [9].

Traditional GMM-HMM systems are constrained by several requirements:

- Handcrafted feature engineering (e.g., delta features).
- Strong assumptions about data distribution (Gaussian mixtures).

- Explicit state duration modeling.

Deep models address these limitations by:

1. **Automating Feature Abstraction:** Learning latent representations through stacked layers (e.g., CNN kernels detecting formants; RNNs capturing temporal dependencies in *Madd* elongation).
2. **Modeling Complex Distributions:** Using non-linear activations to represent intricate spectral patterns (e.g., distinguishing *Qalqalah* echoes from background noise).
3. **Joint Temporal-Spectral Modeling:** Capturing long-range context (e.g., self-attention in Transformers for cross-word *Idgham* rules).

This foundational understanding of *Tajweed* principles and automatic speech recognition (ASR) technologies is essential for effectively contextualizing subsequent research on automated correction systems for Quranic recitation.

## 3 Literature

This section reviews scholarly works that explore technological support for Quranic recitation, excluding direct correction models. It documents relevant applications and web-based resources, focusing on recitation support frameworks rather than corrective algorithmic models, which will be critically analyzed in the following section.

### 3.1 Scholarly Foundations and Contributions

Some publications do not fit neatly into the subsequent thematic classifications and are therefore discussed in this subsection. Research on technological support for Quranic recitation can be broadly categorized into five primary areas: Foundational Literature, Talaqqi-Inspired Models, Rule-Based Search Engines, Dedicated Databases, and Computational Infrastructure.

#### 3.1.1 Foundational Literature

In recent years, numerous scholarly works have addressed Quranic recitation technology, reflecting the ongoing development of the field. This subsection examines selected foundational research relevant to recitation frameworks.

One study [17] reviews literature across three thematic areas: Existing Surveys, State-of-the-Art Papers, and Challenges with their proposed Solutions. However, the review does not cohesively integrate these aspects, resulting in disjointed coverage that fails to explore the relationship between identified challenges and the capabilities or limitations of existing approaches. A more rigorous methodology would involve synthesizing the challenges directly with state-of-the-art research and survey findings to systematically investigate the underlying reasons for accuracy limitations.

Another work [18] provides well-established tables detailing fundamental Arabic phonemes. The third table specifically introduces core and emerging Automatic Speech Recognition (ASR) terminology, including concepts such as Mel-Frequency Cepstral Coefficients (MFCC) and Vector Quantization. The final table synthesizes performance metrics for various techniques, aggregating data from multiple published studies. The results demonstrate that MFCC analysis, when coupled with an effective classifier, outperforms spectrogram-based approaches.

A further study [19], although primarily focused on Natural Language Processing (NLP), appropriately encompasses speech processing as a subdomain. The paper examines approximately 17 scholarly works within this scope, including the specific recitation-related issues addressed by each study and the corresponding techniques employed for resolution. The analysis reveals that the majority of papers utilized Automatic Speech Recognition (ASR) methodologies, while a subset leveraged established tools such as the CMU Sphinx toolkit. Overall, the literature's focus is broader and more generic than the specific objectives of the current research, yet it constitutes a comprehensive synthesis of prior work.

Another review [20] comprehensively examines approximately 12 studies specific to Quranic recitation and 13 pertaining to Arabic speech processing more broadly. However, the research primarily functions as a descriptive compilation rather than presenting novel results. Crucially, the authors aggregate publications without engaging in substantive critical analysis of the methodologies or findings. This approach results in a collection lacking a clear analytical purpose or synthesized insight. Nevertheless, the paper merits acknowledgment for its selection of visually effective and well-presented figures of articulation (Figure 7).

A comprehensive study [21] organizes technological approaches to Quranic recitation into four categories: knowledge-based systems (with correction capabilities), phonetic analysis using spectrograms (incorporating correction), recitation recognition systems (with varying recognition capabilities), and interactive applications (primarily knowledge-focused).

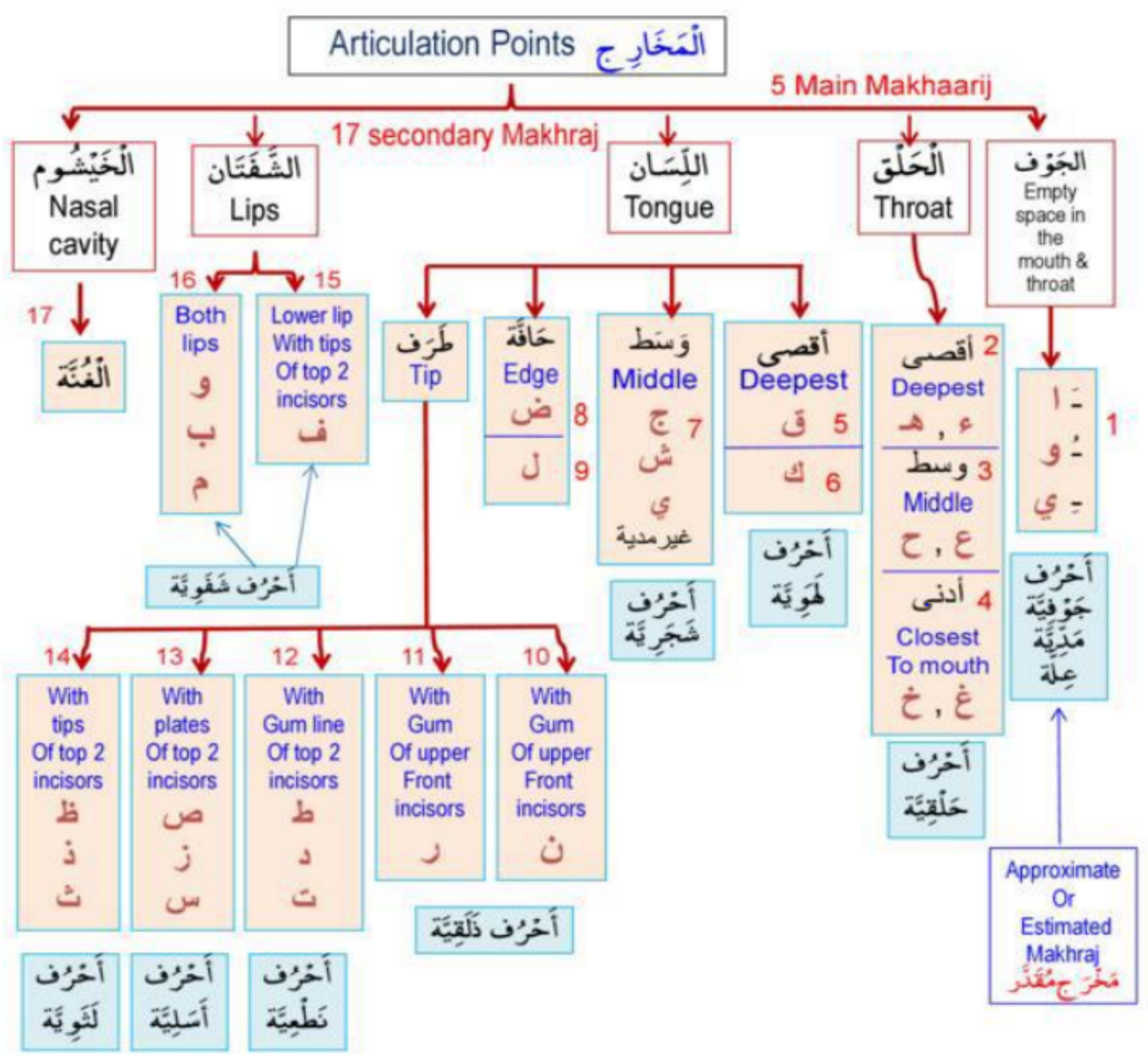

Figure 7: [20] figure of articulation

Though published seven years ago, this work identifies five essential conditions for ethically and technically sound technological approaches to Quranic recitation:

- Technology must comply with Islamic guidelines and avoid spiritual or legal harm
- Technology should serve only as an assisting tool in teaching and learning
- Traditional *Talaqqi Musyafahah* must remain the primary learning method
- All technology must align with Quranic recitation pillars including Arabic language methods, *Rasm Uthmaniy*, and approved *Sanad*
- Development requires collaboration between Quranic scholars and technology experts

The first three conditions are considered essential prerequisites, while the last two serve as recommended enhancements for practical implementation.

Another work [22] presents a conceptual framework for recitation verification without model implementation. Its literature review demonstrates exceptional clarity, particularly in the speech recognition subsection, surpassing many contemporary publications. The proposed model has either been directly implemented or significantly influenced subsequent research using similar architectural principles.

The most notable contribution in this area [9] features an exemplary structure with numerous illustrative figures. It provides detailed explanations of complex ASR algorithms like the greedy decoder, though these technical details fall outside our current scope. Table 4 presents six critical research questions about technological Quranic recitation, each addressed with comprehensive analysis. The paper also proposes a novel model to address limitations in prior research and effectively outlines significant open research challenges.

A rigorous comparative analysis [23] identifies five persistent challenges in Quranic recitation technology:

- Lack of comprehensive datasets
- Need to minimize unacceptable recitation errors
- Diversity of narrations (*Qira'at*)
- Variety in recitation melodies (*Magam*)
- Variations in prolongation (*Madd*) lengths

### 3.2 Methodological and Technical Advances

An enhanced method for Quranic audio segmentation [24] demonstrates significant performance improvements (from 16% to 70%), enabling generation of high-quality datasets for training and evaluation. While algorithmic details are beyond this review's scope, the authors acknowledge room for further refinement.

A study on MFCC parameterization [25] evaluates different filterbank configurations (12, 22, 32, and 42 filters), finding 32 filters optimal at 85% efficiency. The Hamming window function achieved 75% efficiency in the 32-filter setup. More recent studies [26, 27] examine MFCC filter efficacy across distinct *Tajweed* rules and syllables, to be discussed in the following section.

### 3.3 Databases

The QDAT database [28] features crowd-sourced recordings of a single Quranic verse with both correct and intentionally erroneous recitations. Compiled through social media, it contains 1,509 audio files with female contributors outnumbering males by over 3:1.

Another crowd-sourced dataset [29] contains over 7,000 audio files from 1,287 participants. Analysis shows:

- Male contributors slightly outnumber females
- 41% of recordings from individuals aged 14-18 years
- Only 75% of submissions fully adhered to protocol

A computational ontology [30] standardizes *Tajweed* rules with 94% validation accuracy. However, ontological frameworks require perfect precision due to:

- Risk of error propagation in downstream applications
- Theological requirement for infallible reference standards

The Tarteel AI database [31] contains over 25,000 audio recordings from more than 1,200 participants, using a standardized schema for training and evaluating machine learning models.

### 3.4 Informatic Approaches

1. **E-learning platforms** leverage technology to enhance traditional pedagogical approaches:
    - Simulation of *Halaqa* group sessions [32]
    - Child-oriented *Tajweed* platform [33]
    - Pedagogical framework for Malaysian learners [34]
    - Intelligent Tutoring System for *Tajweed* rules [35]
    - MPQu-Berkah audio-visual tool evaluation [36]
    - Mobile application for *Tajweed* education [37]
    - Tarteel.AI architecture [38]
2. **Reading Assistant** research includes an AR-based solution [39] for users without access to physical *Tajweed*-marked Qurans, offering potential for diverse recitation styles and personalization.
3. **Search Engine Systems** include:
    - Question-answering framework for *Tajweed* rules [40]
    - Rule-extraction algorithm for Surah Al-Shu'ara [41]

### 3.5 Web-Based Resources

- **Awesome Arabic-NLP** [42]: GitHub repository of NLP resources for Arabic text processing
- **EveryAyah** [43]: Multimodal Quranic research archive with:
  - Verse-by-verse MP3 recordings
  - High-resolution Quranic text images
  - Timing files and XML datasets
  - Developer tools
- **About Tajweed** [44]: Educational resource explaining *Tajweed* principles with focus on non-Arab learners
- **Quran.com** [45]: Platform for reading, searching, and studying the Quran with translations and recitations
- **Quran Tracker** [46]: Progress tracking template for recitation, memorization, and Tafsir
- **Qari.Live** [47]: Online tutoring platform combining traditional pedagogy with digital delivery
- **Qutor** [48]: Platform connecting learners with certified Quranic recitation experts
- **Quran Guest** [49]: Home-based Quranic tutoring service
- **Quranic Mind Academy** [50]: Premium tutoring with Al-Azhar-certified teachers
- **Al Quran Lab** [51]: Globally accessible Islamic academy
- **Hidayah Network** [52]: Geolocated tutoring with native Egyptian instructors
- **Al Quran Companion** [53]: Accessibility-focused tutoring platform
- **Understand Quran** [54]: Legacy institution with technology-enhanced teaching
- **Quran Guides** [55]: Nonprofit educational platform for structured Quranic instruction

Table 5 summarizes these web-based resources.

### 3.6 Applied Applications and Platforms

Over the past decade, a significant proliferation of applications and platforms dedicated to Quranic recitation has occurred. Analysis of these tools reveals that existing systems can be categorized into the following functional types:

- **Practice Facilitation:** Tools enabling users to recite alongside pre-recorded recitations
- **Online Tutoring:** Platforms utilizing video conferencing for live, remote instruction
- **Tajweed Rule Libraries:** Collections of audio examples demonstrating individual Tajweed rules
- **Digital Resources:** Provision of theoretical Tajweed knowledge through downloadable documents
- **Automated Analysis:** Systems employing AI-driven audio processing for Tajweed evaluation

#### 3.6.1 Digital Musyafahah (Talaqqi) Platforms

Digital *talaqqi* platforms represent online adaptations of the traditional *musyafahah* method, where knowledge is transmitted directly from a qualified teacher to a student. These platforms essentially function as dedicated video conferencing environments for remote Quranic learning. While leveraging established communication technology rather than introducing novel computational breakthroughs, this approach remains widely regarded as the most reliable method for learning *Tilawah* due to direct oversight by qualified instructors possessing formal authorization (*ijazah*).

Representative examples of such platforms include:

- Bayyinah TV (BTV) [56]
- Al Quran Companion [53]
- Understand Quran Academy [57]
- Quran Explorer [58]
- Quran University [59]
- Quran Majeed [60]

| Website | Resource Type | Unique Feature |
|---|---|---|
| Awesome Arabic-NLP [42] | NLP Repository | Comprehensive Arabic NLP resources (papers, tools, datasets) |
| EveryAyah [43] | Research Archive | Multimodal repository for computational analysis |
| About Tajweed [44] | Educational Reference | *Tajwīd* knowledge base with Q&A for non-Arabic speakers |
| Quran.com [45] | Quranic Platform | Multilingual access with translations and word-by-word study |
| Quran Tracker (Notion) [46] | Productivity Tool | Progress tracker for recitation/memorization/*tafsīr* |
| Qari.Live [47] | E-Learning Platform | Tailored *Tajwīd*/*Tafsīr* learning paths |
| Qutor [48] | Tutoring Service | Largest global platform (1k+ tutors) for *Tajwīd*/*Hifz* |
| Quran Guest [49] | Tutoring Service | Home-based Quranic education emphasizing convenience |
| Quranic Mind Academy [50] | Tutoring Service | Al-Azhar-certified teachers with family pricing |
| Al Quran Lab [51] | Tutoring Service | Pakistan-registered non-partisan instruction |
| Hidayah Network [52] | Tutoring Service | Native Egyptian tutors with stringent selection |
| Al Quran Companion [53] | Tutoring Service | Affordable spiritual pedagogy emphasizing *Tarteel* |
| Understand Quran [54] | Tutoring Service | Teacher-led youth programs since 1998 |
| Quran Guides [55] | Tutoring Service | Elite teachers with profit reinvestment |

Table 5: Web-based resources for Quranic studies and recitation

### 3.6.2 Shadowing and Mimetic Practice Platforms

Digital recording technology enables users to practice and self-assess their Quranic recitation by capturing their own recitations. Several platforms leverage this functionality, allowing learners to recite immediately following a pre-recorded expert recitation, listen to their own attempt, and identify areas for improvement through comparative revision.

Representative examples of platforms implementing this recite-after methodology include:

- Al Quran ul-Kareem [61]
- Learn Quran Tajwid [62]
- Quran Majeed [56]
- House of Quran [63]

### 3.6.3 AI-Driven Audio Analysis and Correction Platforms

While numerous platforms are marketed as AI-driven solutions, a critical assessment identifies four applications that demonstrably incorporate substantive audio signal processing techniques combined with machine learning models for Tajweed analysis. These platforms represent the current forefront of automated recitation evaluation technology.

- **Qara'a: Learn Quran** [64] employs a gamified, Duolingo-inspired model for acquiring Quranic recitation, with its core paid feature being the *Murajaah* automated error detection system. However, empirical evaluation reveals significant reliability issues, with the system frequently generating both false positives and false negatives. The system's behavior suggests a reliance on a deterministic rule-based system rather than sophisticated acoustic modeling. Consequently, while Qara'a may serve as a valuable introductory tool for beginners, its technical shortcomings pose a notable pedagogical risk of reinforcing incorrect pronunciation and Tajweed rules.

- **TajweedMate** [65] employs a structured, module-based pedagogy for individual Tajweed rules, complemented by theoretical and audio reference material. Despite this organized framework, its core automated evaluation system demonstrates critically low accuracy. Empirical analysis suggests the model is likely overfit to the specific acoustic profile of its reference reciter, causing it to misinterpret natural vocal variations as errors while failing to robustly identify genuine phonological and Tajweed mistakes. This technical deficiency is compounded by an unclear feedback mechanism that fails to guide learners effectively, posing a significant pedagogical risk of promoting vocal mimicry over correct Tajweed.
- **Tarteel.AI** [66] demonstrates significant limitations in its precision for domain-specific Quranic recitation assessment. The system shows a fundamental inability to process errors in diacritical marks (*Tashkeel*) and minimal capability to detect established Tajweed rule violations. Most critically, it frequently misclassifies correct application of rules like phonetic elongation (*Madd*) as errors. These observations indicate the underlying model is optimized for generic Arabic speech recognition rather than specialized Tajweed evaluation, posing a risk of providing incorrect feedback to users.
- **Al Siraat** [67] features a theoretically sound pedagogical structure that separates foundational pronunciation from full-verse recitation. However, the early-access version is critically hampered by implementation failures, including fundamental inaccuracies in its core pronunciation module, systemic technical instability due to server dependencies, and unresolved data privacy concerns from mandatory off-device audio processing. Compounding these issues are transparency deficits, exemplified by the use of AI-generated avatars in lieu of verifiable developer credentials. Consequently, the platform's severe technical flaws and governance questions currently invalidate its theoretical benefits.

The summary of the evaluation is shown in Table 6.

## 4 State of the Art

Over the past decade, research in computer science has increasingly focused on improving Quranic recitation technologies. This section reviews and critically evaluates key contributions in this domain.

The system presented in [68] suffers from fundamental limitations due to its training data, which relies on experts simulating errors rather than authentic learner recordings. This approach introduces artificial error patterns that lack natural variation, thereby limiting the model's diagnostic capabilities. The arbitrary selection of error variants prevents specific feedback generation, while the absence of female recitations in the training data introduces gender bias and compromises generalizability. While a cohort of ten reciters might appear sufficient, validation with genuine learner data remains essential to address these constraints.

A subsequent study [69] claims to achieve 97.7% accuracy, yet retains critical limitations. The verification process focuses on only a narrow subset of Tajweed rules, and the training data lacks gender diversity. These unaddressed issues in scalability and bias significantly constrain the model's applicability for comprehensive recitation assessment.

The methodology in [70] is critically flawed by its data trimming process, which risks removing essential Tajweed characteristics such as *Hams* and introduces artificial audio boundaries. This causes the model to learn extraneous features rather than contextual rules. Additionally, the system demonstrates pronounced gender bias against female reciters due to insufficient pitch variation in the training data and provides only word-level accuracy without diagnostic error feedback. The approach's reliance on extensively processed data presents significant challenges to practical scalability and real-world implementation.

The lip-movement analysis approach in [71] is fundamentally misaligned with Quranic phonetics, as it can only assess a limited subset of articulations related to *Tashkeel* rather than the full range of *Makharij Al-Huruf*. The system's practical utility is further limited by demanding hardware requirements and the need for real-time, robust visual feedback to be pedagogically effective. Despite these limitations, the paper contributes a unique, non-ASR approach to the field.

The LSTM-based approach in [72] reports 96% accuracy in modeling temporal dependencies, but its scope is critically limited to only three Tajweed rules. This raises concerns about scalability to the Quran's full phonetic complexity. The evaluation on a single verse fails to validate applicability across the text's 6,000+ verses, and performance lacks benchmarking against expert human evaluation. The model's robustness remains unproven for verses with multiple or overlapping rule instances, as the QDAT dataset structure did not test this scalability challenge.

The architectural shift to an EfficientNet-B0 backbone with attention mechanism in [73] claims only a superficial 2% accuracy improvement while failing to address the core methodological flaw of its predecessor: a binary classifier evaluating entire verses. This design cannot localize errors to specific characters or words when multiple rule instances occur within a single verse, rendering its feedback diagnostically ineffective. The absence of real-time performance validation under practical deployment scenarios further undermines claims of system viability for interactive learning.

| Criteria | Qara'a [64] | TajweedMate [65] | Tarteel.AI [66] | Al Siraat [67] |
|---|---|---|---|---|
| Core Technology | Gamified learning + Error detection | Rule-modules + Recitation evaluation | Arabic speech recognition | Word-level + Verse-level processing |
| Accuracy/Reliability | Low reliability (High false positives/negatives) | Very low accuracy (Fails basic mimic tests) | Incorrect Tajweed assessment (High false positives on correct recitation) | Non-functional in current state |
| Key Technical Shortcomings | • Likely rule-based lookup table<br>• Minimal acoustic pattern recognition<br>• Suspected low-quality training data | • Severe overfitting to reference reciter's voice features<br>• Unclear error feedback mechanism | • Cannot detect *Tashkeel* errors<br>• Misclassifies *Madd* as errors<br>• Generic speech model $\neq$ Tajweed evaluation | • Phonological errors in training data<br>• Server dependency causing failures<br>• Unverified AI-generated promotion |
| Pedagogical Value | • Useful introductory tool for beginners<br>• Risk of reinforcing incorrect pronunciation | • Well-structured theory content<br>• Promotes harmful vocal mimicry over correct Tajweed | • Misleading for learners<br>• Penalizes correct recitation | • Theoretically sound structure<br>• Currently unusable for learning |
| Data Handling | Not specified | Not specified | Not specified | Off-device processing with security concerns |
| Research Viability | Limited (Flawed core methodology) | Low (Fundamental model flaws) | Counterproductive (Misaligned objectives) | Not viable (Pre-release instability) |

Table 6: Comparison of AI-Driven Quranic Recitation Applications

The foundational system in [74] (Figure 8) is critically limited by its likely localized dataset, introducing potential socio-linguistic bias and lacking transparency. Its reported performance is compromised by a 17% false acceptance rate and a hypothesis generation module that correctly identifies actual errors only 52.2% of the time. The system's lack of real-world adoption over two decades raises significant doubts about its practical feasibility and scalability.

The approach in [75] demonstrates performance gains through Speaker Adaptive Training (SAT) but suffers from a fundamental conceptual misalignment. It conflates phonetic units with the nuanced, context-dependent rules of Tajweed, ignoring critical suprasegmental features and rule interactions. A significant validation mismatch persists, as the system was tested only on native Arab reciters despite targeting non-native speakers, rendering its performance metrics unreliable for the intended user base.

The refinements in [76] show slight improvements in phone recognition and reduced user calibration time through SAT, though core phone recognition performance remained stable at 90%. However, the research is limited by insufficient evaluation metrics, unresolved audio segmentation issues, and inadequate validation against human expert judgments.

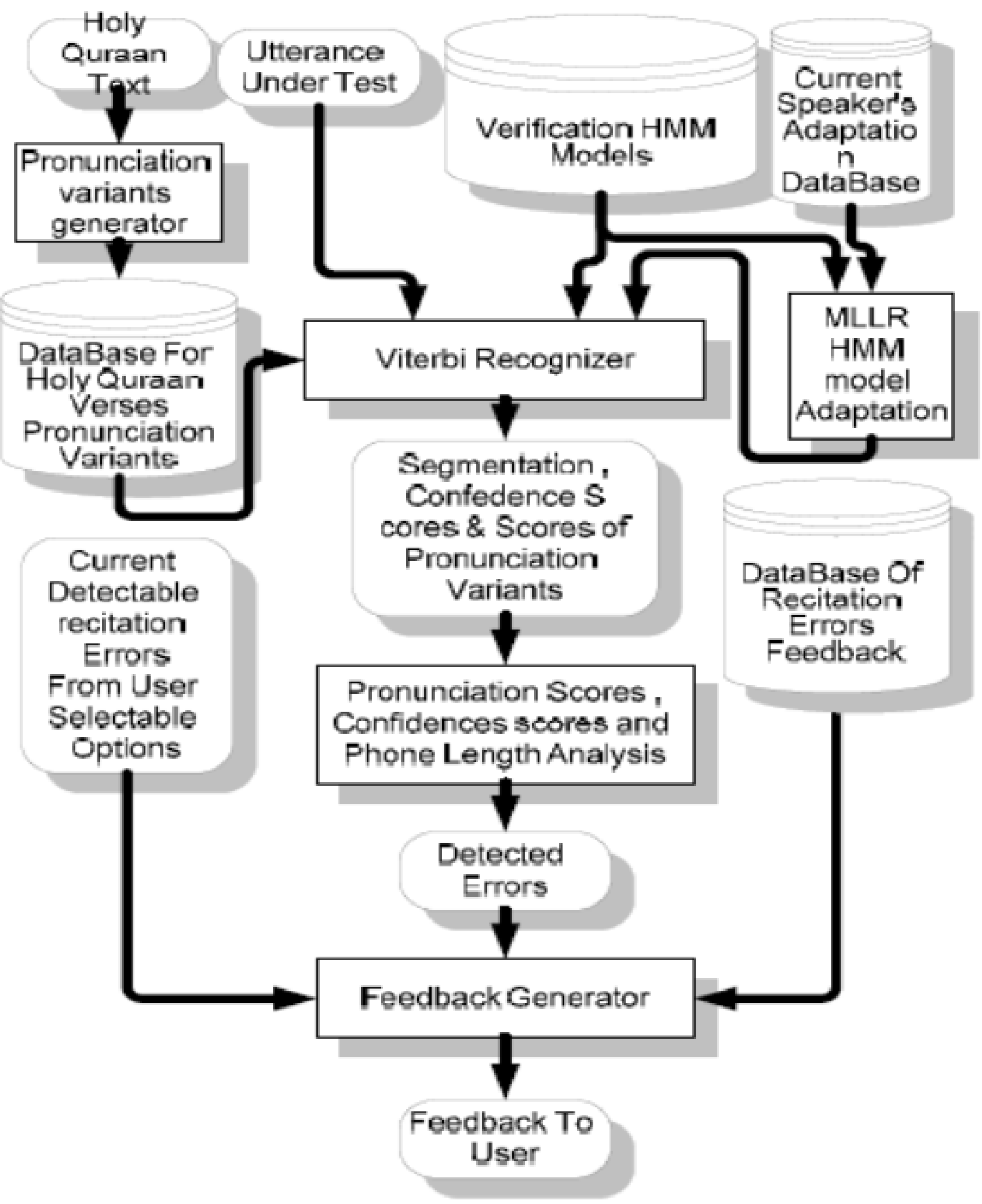

Figure 8: [74]'s model architecture (HAFSS)

These methodological shortcomings preclude robust assessment of the HAFSS system's (Figure 9) true efficacy for Tajweed evaluation.

The work in [77] reports performance gains for the HAFSS system but adopts a publication style emphasizing technical description over rigorous empirical validation. The experimental design contains critical flaws, including allowing HAFSS users extra time, which introduces confounding variables. The system's 23.9% repeat request rate, 4.1% expert disagreement rate, and unaddressed diagnostic inaccuracies necessitate substantial methodological refinement before it can be considered robust or pedagogically reliable.

The hybrid HMM-DNN modeling approach in [78] using MFCC features reports a marginal 1.02 percentage point improvement over its best HMM configuration to reach 91.5% accuracy. While demonstrating enhancement relative to the HAFSS benchmark, this gain remains modest. The authors appropriately acknowledge that further methodological enhancements are necessary, as this incremental improvement does not represent a significant advancement for the field.

The system in [80] operates solely as a classifier, identifying only the most probable *Makhraj* region without providing the detailed, diagnostic feedback on articulation precision necessary for effective *Tajweed* learning. Its exclusive focus on isolated characters represents a critical limitation, as it disregards the contextual variations and influence of *Harakah* (diacritical marks) that are essential for correct pronunciation in actual recitation.

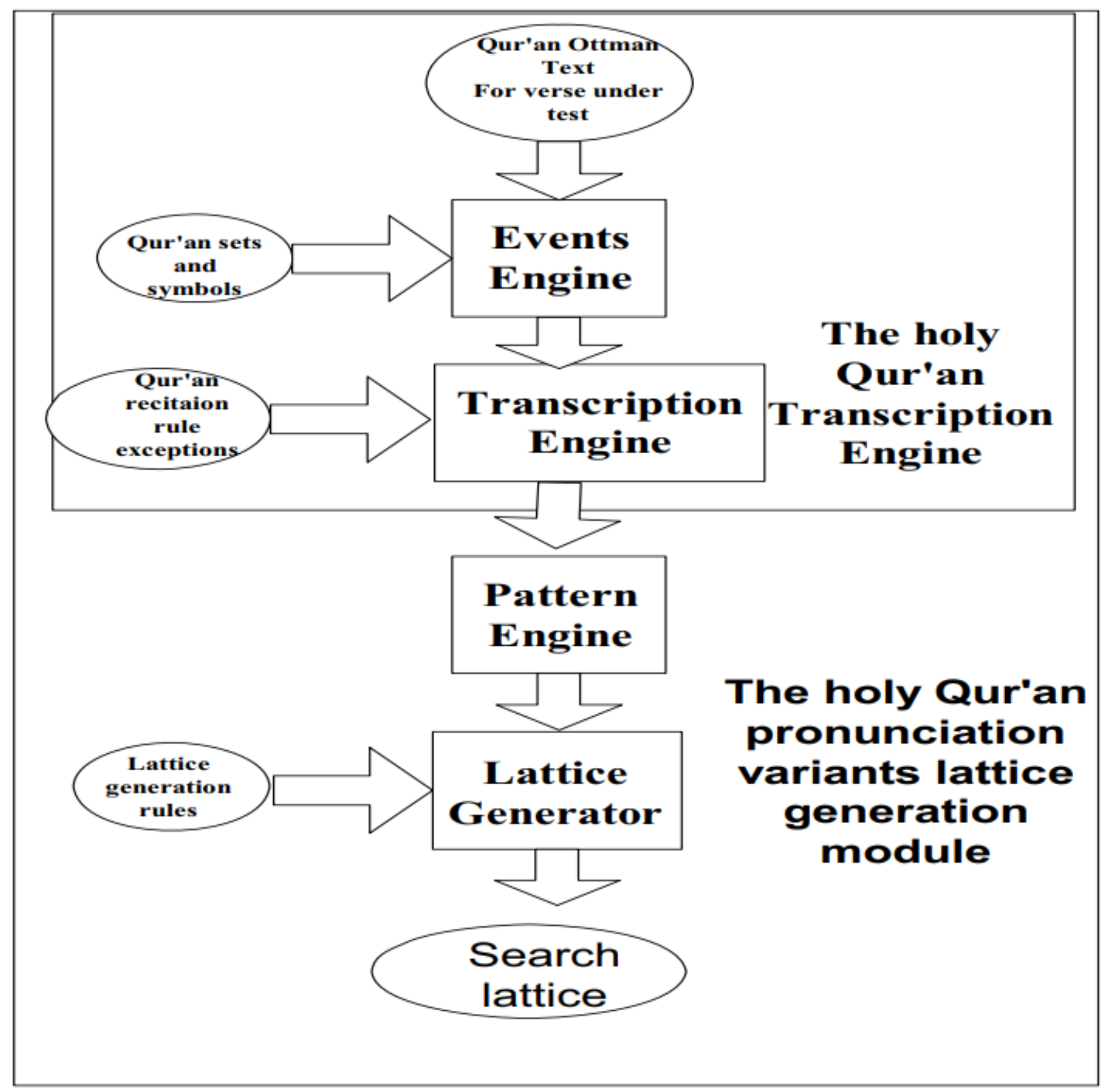

Figure 9: [79] architecture for error hypothesis generator

While [81] enhances the dataset and technical exposition, it retains the same fundamental methodological limitation. The system offers no substantive improvement in pedagogical utility or contextual evaluation, as it continues to lack the granular feedback required for meaningful *Tajweed* instruction.

The methodology in [82] features a sound pedagogical design but is critically undermined by a complete absence of noise suppression. This omission corrupts its MFCC-based feature extraction and invalidates all results. Additionally, its testing protocol lacks reliability, as recitations were not validated by a *Tajweed* expert to establish ground truth. The use of separate GMMs for each rule also incurs unsustainable computational costs. For credible findings, future work must incorporate noise cancellation, expert-validated testing, and an optimized balance between model complexity and computational feasibility.

The pitch-based feature extraction method in [83] demonstrates computational efficiency and thorough preprocessing. However, it exhibits reduced sensitivity, resulting in lower overall accuracy compared to MFCCs and the erroneous flagging of correct recitations. While its high precision and real-time efficiency (documented RTF 90%) present notable practical advantages, its discriminative capability remains a core weakness. Targeted optimization could enhance its utility for speed-critical applications, but its current performance remains inferior to established methods.

The study in [84] exhibits a fundamental incongruity, as its title promises ”correction” while the methodology delivers only recognition and classification without any mechanism for corrective feedback. The research is further compromised by an impractical dataset collected through uncontrolled interfaces, introducing significant variability that undermines model reliability. Most critically, the validity of the entire study is threatened by a lack of transparency in data labeling, risking a feedback loop where potentially flawed algorithmic classifications from software like ”Qara” could define the ground truth, systematically propagating errors.

The work in [85] primarily advocates for MFCCs, representing a methodological approach similar to prior HAFSS work rather than introducing novel techniques. Its subsequent expansion in [86] is fundamentally flawed, demonstrating poor scalability with rising word error rates on longer verses. The system critically misclassifies a *Tashkeel* error as a *Madd* rule violation. Furthermore, its development and evaluation solely on Surah Al-Fatihah—which lacks complex *Tajweed* rules—renders its benchmark unrepresentative and its practical utility for comprehensive recitation severely limited.

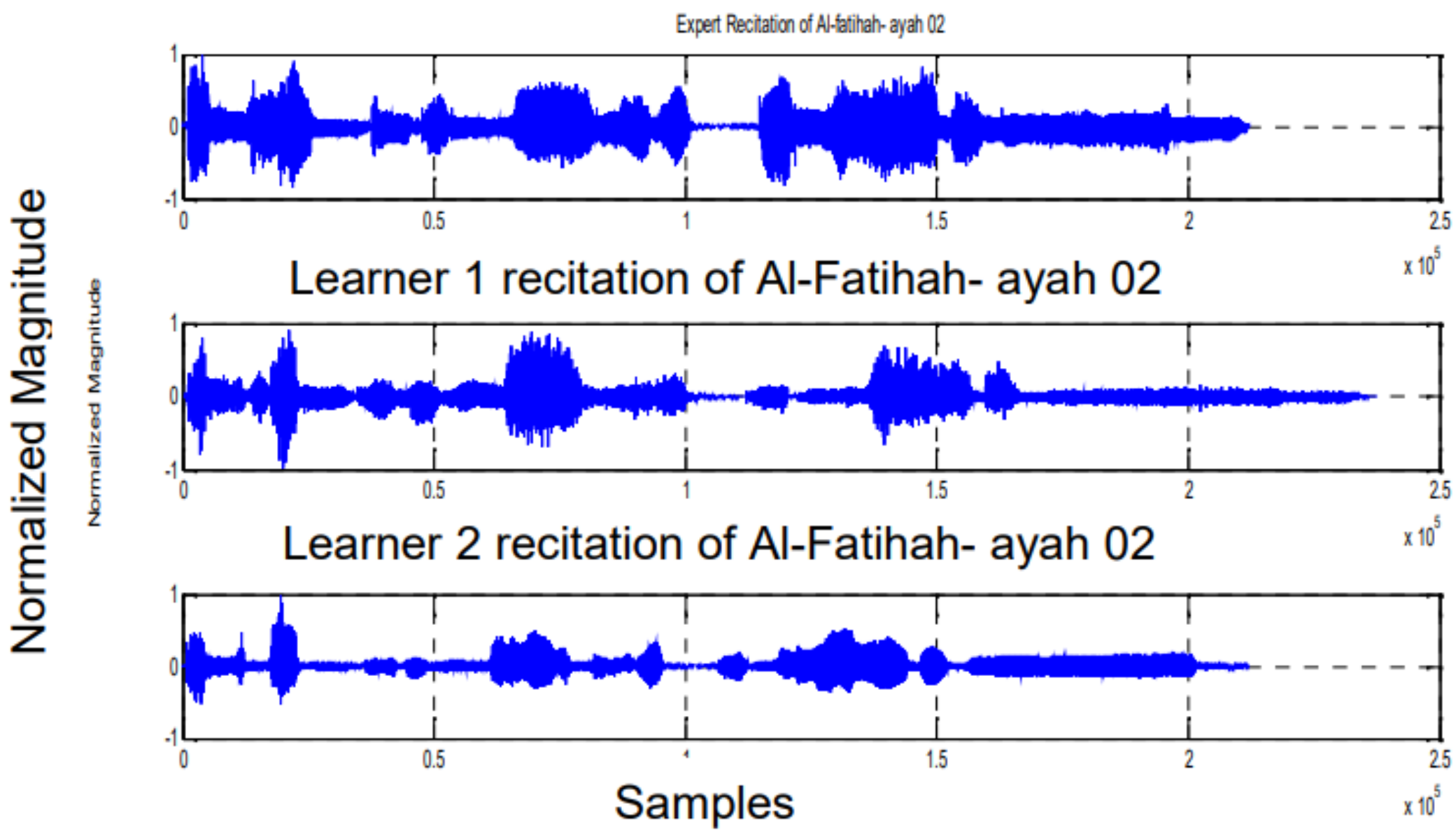

Figure 10: [88], before DTW

The framework in [87] integrates visual guidance with an automated evaluation model that processes audio through feature extraction and classification stages, employing Dynamic Time Warping (DTW) and a threshold-based correction system. However, the model's reliance on a simple correctness threshold per segment lacks the granularity to provide specific diagnostic feedback on the nature of *Tajweed* errors. This design, coupled with an absence of reported validation against expert evaluations or on complex rules, significantly limits its pedagogical utility for precise correction.

In their previous work, the application of Dynamic Time Warping (DTW) in [88] significantly reduces the Mean Squared Error (MSE) between learner and expert frames by normalizing temporal inconsistencies, as shown in Figures 10 and 11. However, this alignment process discards the learner's original temporal performance data, which may contain crucial diagnostic information about their recitation rhythm and pacing. Consequently, while DTW enhances feature extraction consistency, its impact on the accuracy and pedagogical utility of actual error correction remains unresolved, as the system may overlook important temporal aspects of pronunciation.

The Nur-IQRA framework introduced in [89] utilizes MFCC for feature extraction and an HMM-GMM approach for parameter estimation. The model calculates similarity between learner and expert recitations using both a classical fuzzy set methodology and Euclidean distance applied to vowel characteristics. While this combination of techniques aims to refine the evaluation process, the pedagogical efficacy and specific diagnostic capability of these combined metrics for *Tajweed* error correction remain to be fully validated.

The IQRA system evaluated in [26] employs a sophisticated three-stage pipeline incorporating Vocal Tract Length Normalization (VTLN) and Dynamic Time Warping (DTW). However, it suffers from significant accuracy degradation, with a cripplingly high False Rejection Rate (FRR) of 81.98%. This poor performance, yielding observed accuracies between 82-87%, is attributed to a switch from HMM to an MLE-based classifier and a speculative band-based MFCC categorization (Table 7) that hypothesizes but does not validate a correlation with *Tajweed* rules. The system's fundamental failure lies in its substantial classification inconsistency, rendering it unreliable for practical pedagogical application.

The study in [27] enhances classification by replacing a threshold system with a sequential ensemble of Linear Discriminant (LD), Support Vector Machine (SVM), and K-Nearest Neighbors (KNN) classifiers. The KNN classifier achieved 92.24% accuracy using spectral band 1 features. Crucially, the study found that optimal performance requires band-specific feature selection tailored to distinct error categories (*Makhraj*, *Tajweed*, *Sifat*, *Harakaat*), rather than universally applying a single band. This finding demonstrates that the choice of MFCC spectral band is critically dependent on the specific type of recitation error being classified, as illustrated in Figure 12.

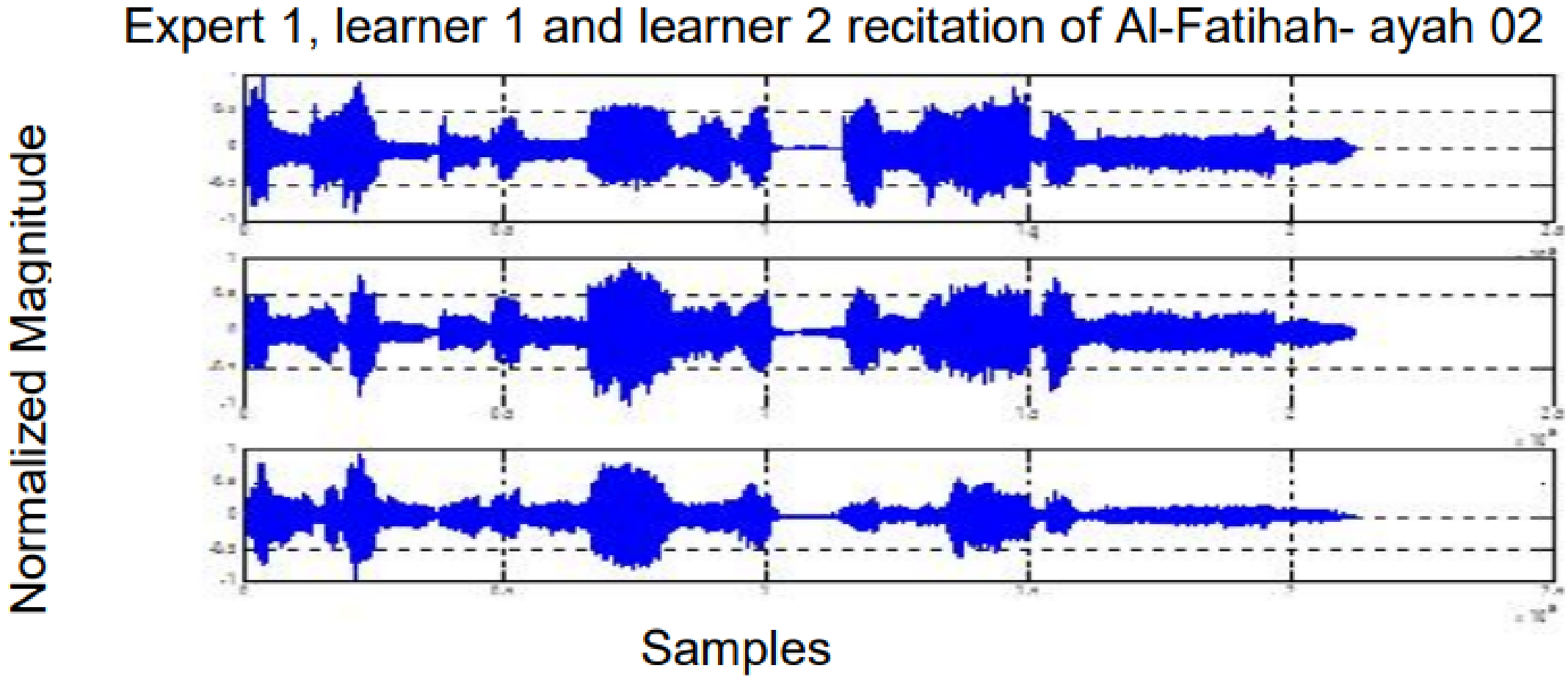

Figure 11: [88], after DTW

| Fragmentary Frequency | Band1 (300Hz-1077Hz) | Band2 (861.3Hz-2089Hz) | Band3 (1787Hz-3704Hz) |
| --- | --- | --- | --- |
| Syllable Pronunciation | High-Voiced Vowel and Low-Voiced Consonant | High-Voiced Consonant and Low-Voiced Vowel | Voiceless and Low-Voiced Consonant |
| MFCC-Formant co-efficient | C1, C2, C3 and C4 | C5, C6, C7 and C8 | C9, C10, C11 and C12 |

Table 7: [26]'s MFCC filters bands

The comprehensive system in [90] reports achieving 87.27% accuracy on Surah Al-Fatihah, representing a notable decrease from the 92.24% accuracy of its standalone KNN classifier. This performance drop highlights the integration cost within a full processing pipeline. The decline is likely attributable to the system's sequential processing stages and the distinct linguistic profile of Surah Al-Fatihah, which emphasizes *Makhraj* and *Madd* over complex *Tajweed* rules. Furthermore, the thesis's methodological details remain largely inaccessible, as its findings are derived from limited public information rather than a complete technical exposition.

The study in [91] proposes an optimized Dynamic Time Warping (DTW) and Discrete Wavelet Transform (DWT) methodology for *harakat* analysis. However, the system achieves only marginal accuracy ( 80%) and produces critically invalid outputs, such as mapping a 45ms duration to an invalid 3-*harakat* category. This failure stems from an inability to model natural temporal variations in recitation speed, causing pronounced errors on specific syllables and high False Rejection Rates (FRR). Consequently, the system's performance is undermined by unaddressed pedagogical and technical challenges, limiting its utility for reliable vowel duration assessment.

The duration-based model for *Tajweed* rules proposed in [92] suffers from accuracy issues due to its dependence on subjective duration values (e.g., 2, 4, or 6 counts, depending on *Maad*) assigned by experts during data annotation. This subjectivity leads to potential misclassification of permissible recitations. Furthermore, the threshold-based approach fails to adequately handle categorically incorrect intermediate durations, such as 3 counts, which fall between permissible values but are themselves invalid. This methodological flaw introduces significant inconsistency and limits the model's reliability for rules with inherent variability.

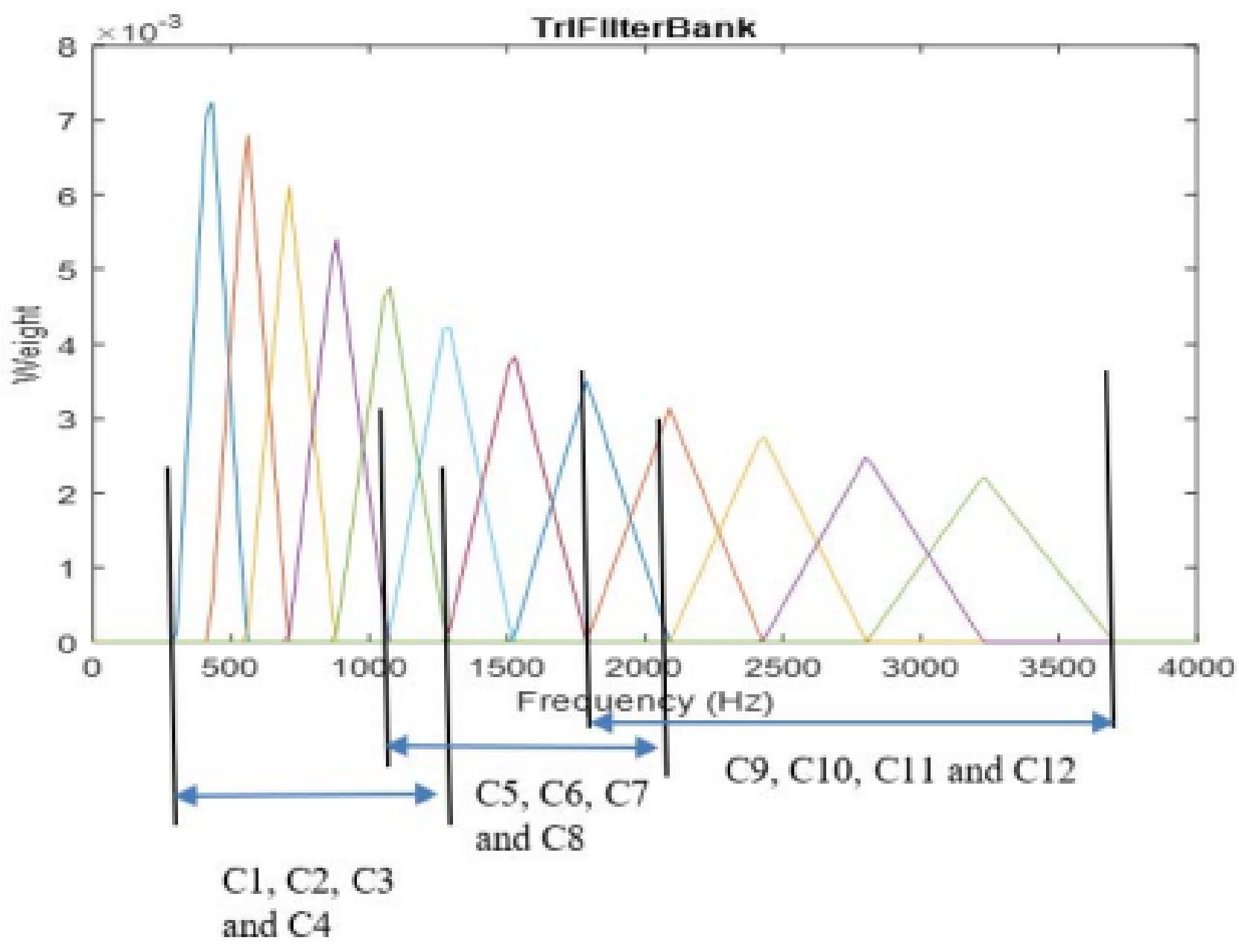

Figure 12: [27]'s MFCC bands in frequency diagram

The work in [93] is severely limited by a non-generalizable dataset from only two individuals and unaddressed audio artifacts like echo, which corrupt the input data. The suboptimal accuracy of 77.7%, achieved with a Levenberg-Marquardt algorithm, provides only a weak indication of the potential efficacy of Artificial Neural Networks (ANNs) for this task. Consequently, the study's findings lack validity and require a more rigorous approach with a larger cohort of experts and robust audio preprocessing to be credible.

The study by [94] reports a high accuracy of 97.3% in recognizing the *Makhraj* of the first seven Arabic letters under noisy conditions. However, its scope is critically limited to isolated characters, rendering it ineffective for full-word pronunciation or comprehensive *Tajweed* assessment. The study prioritizes noise suppression techniques like the Normalized Least Mean Squares (NLMS) algorithm, yet this narrow focus on single-character utterances ignores the contextual and continuous nature of actual recitation. Consequently, while a notable technical exercise in noise resilience, the model's design offers negligible utility for practical, pedagogical application.

The model proposed in [95] for isolated *Makhraj* recognition presents a critical methodological flaw by providing no experimental testing or validation of its proposed system. The model is designed solely for single-character recognition (*Hijaiyah*), and while the authors suggest K-Nearest Neighbors (KNN) is theoretically suitable, the complete absence of results prevents any verification of this claim. Consequently, the paper remains a purely theoretical proposal with no demonstrable performance or practical utility.

The work in [96] demonstrates high accuracy on isolated letters using RelAtive SpecTrAl Perceptual Linear Prediction (RASTA-PLP) and Hidden Markov Models (HMMs). However, its results are inflated due to a constrained scope and a lack of rigorous operational definition for "mispronunciation." The model's performance on a limited set of letters fails to demonstrate scalability to continuous speech or the complex co-articulation governed by *Tajweed* rules. Consequently, its reported success is superficial and masks fundamental challenges in generalizability and practical application.

The study in [97] employs Mel-Frequency Cepstral Coefficients (MFCC) and Convolutional Neural Networks (CNN), reporting a 90.8% accuracy for *Tajweed* rule classification. However, this result is critically undermined by testing on the same reciter used for training, introducing severe overfitting and a lack of generalizability. The dataset's exclusion of female reciters introduces demographic bias, and the model fails to address acoustic artifacts like reverberation in web-sourced recordings. Consequently, while technically sound for idealized male recitations, the system's applicability to diverse, real-world recitations remains highly constrained.

Utilizing the Kaldi toolkit (Figure 13), the work in [98] shows a remarkably low Word Error Rate (WER) of 0.27% for Automatic Speech Recognition (ASR) on a specific surah. However, its scalability to the entire Quran is hindered by the substantial data required for its Deep Neural Network (DNN)-based acoustic model to handle the text's full phonological complexity. More critically, the framework is designed for speech recognition, not pedagogical assessment. Adapting it for *Tajweed* verification would impose far stricter precision requirements on phonemic and prosodic accuracy. This shift in objective would likely degrade its performance metrics, as the model lacks the architectural refinements necessary for the sensitive task of distinguishing correct from incorrect recitation.

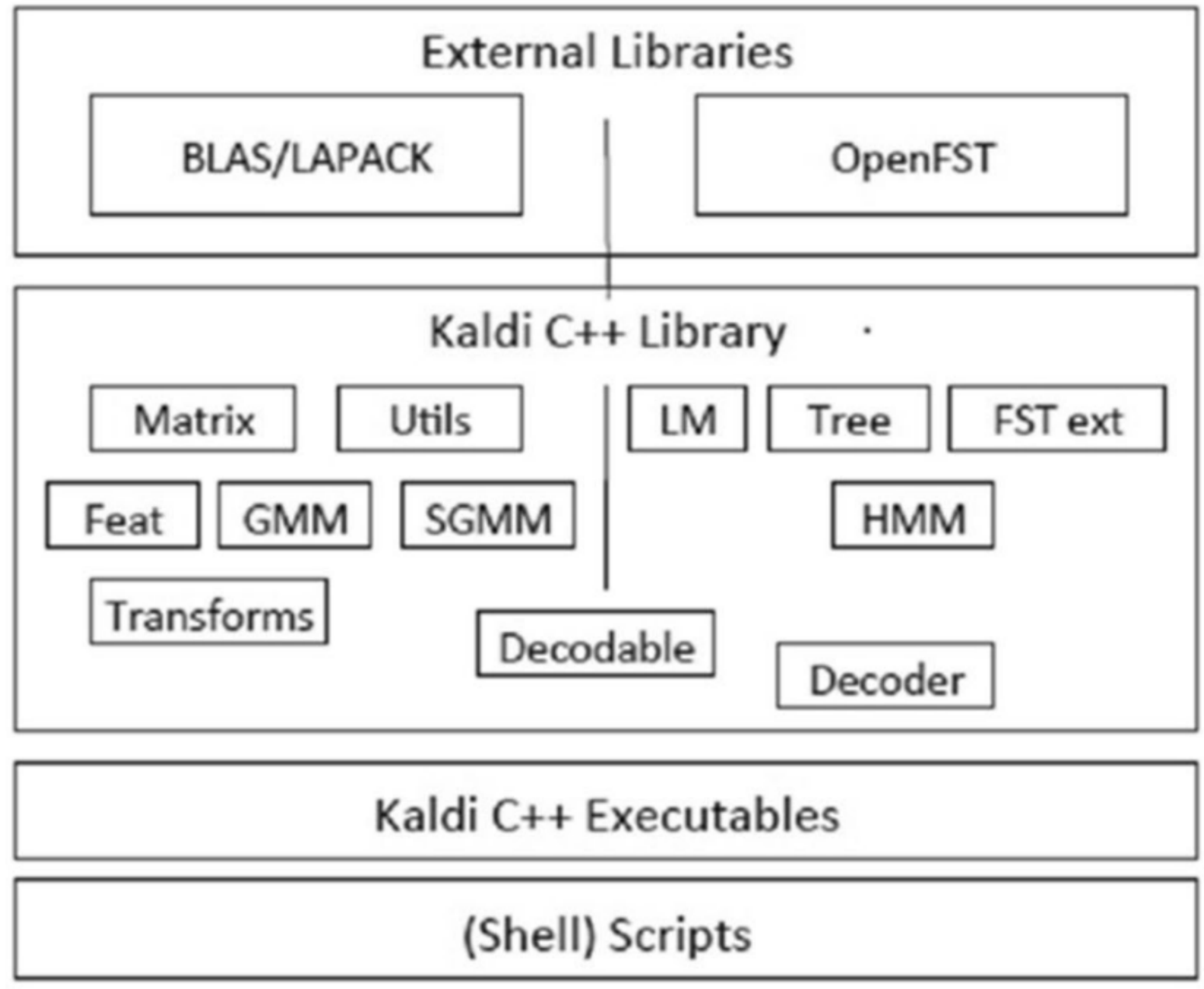

Figure 13: Kaldi toolkit [98]

### 4.1 Recent Advances and Limitations

The study in [99] addresses a critical gap by introducing the QRFAM database to include female reciters. However, experiments reveal severe performance degradation in cross-gender testing, with a high overall Word Error Rate (WER) of 40.6%. This indicates that standard Mel-Frequency Cepstral Coefficients (MFCC) features encode gender-specific acoustic characteristics, fundamentally limiting model robustness across demographics. Consequently, the study highlights a critical challenge in achieving generalizable recitation assessment without gender-adaptive feature extraction or modeling techniques.

The work in [100] implies near-perfect accuracy ( 100%) in classifying *Medd* types using an audio-based Relative Phase Difference and Amplitude Modulation Coefficient (RPDAMC) algorithm and Hidden Markov Toolkits. However, it functions only as a classifier with limited capability for actual error detection or pedagogical correction. While its audio-centric approach holds theoretical advantages over lexicon-dependent models for future error detection, its current utility remains constrained. Furthermore, significant scalability challenges prevent this methodology from being extended to the full spectrum of *Tajweed* rules due to their immense phonological complexity and contextual dependencies.

The system in [101] utilizes the CMU Sphinx toolkit, a technically sound Automatic Speech Recognition (ASR) foundation. However, it fails to adapt to the unique prosodic and phonological conventions of Quranic recitation. The framework inadequately represents Arabic diacritics and cannot accommodate the rule variability inherent in different recitational traditions (*qira'at*), such as context-dependent prolongation (*maad*) rules. Consequently, while the core technology is appropriate, the implementation lacks the linguistic and recitational complexity required for robust *Tajweed* verification.

The study in [102] claims exceptional accuracy ( 99%) for four *Tajweed* rules using filter banks and Support Vector Machines (SVM). However, its scope is critically limited to this small subset, with performance likely degrading on more complex rules. Furthermore, the architecture is inefficient for holistic assessment, as it processes only one rule per verse, creating scalability issues for verses with multiple co-occurring rules. While a significant technical advancement, the system requires substantial architectural refinement to accommodate the full spectrum of *Tajweed* rules efficiently.

The work in [103] demonstrates promising but non-generalizable results, as its models suffer from significant accuracy degradation on unseen verses due to overfitting on full spectrograms and a critical omission of *Ghunnah* segmentation. The study's use of a non-expert-validated QDAT variant and lack of dimensionality reduction (e.g., MFCCs) further limit its practical utility. Consequently, despite testing advanced architectures like Convolutional Neural Networks (CNN) and Recurrent Neural Networks (RNN), the models lack the representational and preprocessing refinements needed for robust, real-world *Tajweed* assessment.

The study in [104] reports high accuracy ( 98%) using MFCC and Long Short-Term Memory (LSTM) networks. However, its validity is critically undermined by an extremely narrow dataset from a single short surah annotated by one expert, resulting in reciter-specific bias and limited rule coverage. The system's operational paradigm is pedagogically misaligned, as it paradoxically requires expert-level recitation to generate feedback, contradicting its purpose for learner correction. Consequently, while the LSTM architecture shows promise, these methodological flaws prevent the framework from achieving generalized *Tajweed* proficiency or practical utility.

The system in [105] employs a Continuous Hidden Markov Model-Gaussian Mixture Model (CHMM-GMM) ASR using the CMU Sphinx framework with a custom phonetic inventory that contains substantial phonological omissions, fundamentally limiting its utility for *Tajweed*. The system is further compromised by training on commercially processed recitations, which risks encoding artificial enhancements as normative features, and by evaluating only lexical accuracy rather than *Tajweed*-specific rules. Consequently, despite its technical sophistication, the framework operates as a generic Arabic speech recognizer, suffering from domain misalignment and suboptimal performance for authentic Quranic recitation assessment.

The work in [106] reports high accuracy using MFCCs and threshold-based classification. However, its claims are undermined by a critical lack of transparency in how pronunciation ground truth was validated by experts. The framework's applicability is severely limited by its narrow phonological scope, as it was only validated on a small subset of phonemes with no demonstration of scalability to the full *Tajweed* rule system. While methodologically innovative for its time, the approach requires extensive recalibration and rigorous expert validation to be viable for comprehensive *Tajweed* verification.

The study in [107] establishes a methodologically rigorous Computer-Assisted Language Learning (CALL) framework by innovatively rejecting conventional Hidden Markov Model (HMM) classifiers in favor of a hybrid HMM-recognition and discriminative-classifier architecture (T, R classifiers) for phone-level verification. This approach demonstrates 91.2% word-level accuracy by effectively addressing HMM limitations in spectral similarity and error detection, though it intentionally excludes certain *Tajweed* rules like *Maad*. Consequently, it provides a seminal, high-precision foundation for *Makhraj* analysis while acknowledging its deliberately limited scope within the broader *Tajweed* system.

The work in [108] reports high accuracy (97.5%) for isolated word classification using a Deep Neural Network (DNN)-MFCC architecture. However, its utility is critically limited by a whole-word approach that provides no granular error localization and inherently neglects inter-word *Tajweed* phenomena like *Idgham*. This design results in a fundamental misalignment with pedagogical objectives, as it offers mere pronunciation assistance rather than holistic recitation assessment. While technically rigorous within its constrained lexicon, the framework requires integration of phone-level diagnostics and cross-word phonological modeling to be viable for comprehensive *Tajweed* training.

## 5 Discussion

The pursuit of automated Quranic recitation evaluation represents a formidable challenge at the intersection of computational linguistics, audio signal processing, and Islamic sacred science. As this review elucidates, the field is characterized by a persistent and critical gap between technological ambition and pedagogical efficacy. The prevailing paradigm, predominantly reliant on adaptations of Automatic Speech Recognition (ASR) architectures, is fundamentally misaligned with the nuanced requirements of *Tajweed* assessment.

This section synthesizes the critical limitations of current approaches, informed by the extensive literature and state-of-the-art analysis, and articulates the rationale for a principled, knowledge-centric alternative. The systemic misalignment manifests in a consistent set of critical failures, as synthesized in Table 8.

| Category | Limitation Description | Related Papers |
|---|---|---|
| *Narrow Objective* | Focuses on phonetic classification rather than holistic evaluation and feedback. | [68], [69], [80], [81], [84], [94], [96], [100], [102], [106] |
| *Limited Rule Set* | Covers only a small subset of *Tajweed* rules, lacking comprehensiveness. | [68], [69], [72], [109], [93], [97], [100], [102], [103], [106], [107], [108] |
| *Focus on Articulation* | Addresses only letter pronunciation (*Makhraj*), neglecting other *Tajweed* rules. | [80], [81], [71], [94], [96] |
| *Small Dataset* | Evaluated on a limited number of Quranic chapters, not the entire Quran. | [85], [27], [109], [90], [98], [101], [104] |
| *Lack of Expert Validation* | Lacks validation by *Tajweed* experts, violating the traditional principle of *Isnad*. | [76], [82], [103], [106], [107], [108] |
| *ASR vs. Evaluation* | Focuses on Automatic Speech Recognition (transcription), not on *Tajweed* evaluation. | [105], [98], [88] |
| *Performance Issues* | Suffers from high computational complexity, long processing times, or low accuracy. | [82], [74], [75], [76], [77], [85], [90], [93], [99], [105] |

Table 8: A categorization of limitations in computational Tajweed research, highlighting persistent challenges across seven key dimensions

The core inadequacy of ASR-based systems stems from their intrinsic objective: transcribing speech into text. While effective for semantic recognition, this objective is orthogonal to the qualitative evaluation of elocution. Acoustic and language models within ASR pipelines are optimized for comprehensibility, often overlooking phonetic inaccuracies to achieve robust word recognition. Consequently, systems like those evaluated in the applications section (e.g., Tarteel.AI, Qara'a) frequently fail to detect subtle yet theologically significant errors in *Tashkeel* (diacritics) or misclassify correct application of rules like *Madd* (elongation) as mistakes. This misalignment renders them pedagogically unreliable and potentially harmful, as they risk reinforcing incorrect pronunciation or penalizing faithful recitation.

A second, profound limitation is the endemic data dependency and inherent biases of data-driven models. As evidenced by the reviewed literature, the performance of Deep Neural Networks (DNNs) and other machine learning models is contingent upon vast quantities of high-quality, authentically erroneous training data. The curation of such datasets, as attempted in [28] and [29], presents immense practical and ethical challenges. Artificially generated errors [68] lack the acoustic authenticity of mistakes made by genuine learners, while crowd-sourced data is often contaminated by inconsistencies and a lack of expert validation [84]. Furthermore, these datasets frequently exhibit severe demographic biases. As demonstrated in [99], models trained primarily on male reciters exhibit catastrophic performance degradation when applied to female or child voices, whose vocal tracts produce different fundamental frequencies and formant structures. This bias systematically excludes a significant portion of the Muslim populace from benefiting from such technologies.

Moreover, the current landscape is fragmented by a narrow focus on isolated sub-problems. Many studies, such as those focusing solely on *Makhraj* classification for isolated letters [80, 96] or a limited subset of *Tajweed* rules [72, 102], report high accuracy on constrained tasks. However, these approaches fail to scale to the continuous, context-dependent nature of full-verse recitation, where rules interact and co-articulation effects are paramount. The pedagogical utility of such systems is further undermined by a pervasive lack of granular feedback. Most models provide a binary correct/incorrect output or an overall score, failing to identify the specific nature and location of an error, which is indispensable for effective learning, such as [87].

These critiques collectively underscore a fundamental paradox: the Quranic text is uniquely immutable and rule-governed, yet most technological approaches treat it as a dynamic, statistical problem. The Quran's text, its canonical recitations (*Qira'at*), and the rules of *Tajweed* are precisely defined and have been preserved for centuries through the rigorous tradition of *Isnad* (chain of transmission). This invariance presents a remarkable opportunity. Instead of relying on statistical patterns learned from imperfect data, a more robust system can be architected around *a priori* knowledge.

## 6 Conclusion

This review has traversed the extensive landscape of technological interventions aimed at automating the evaluation of Quranic recitation. The analysis culminates in a clear and decisive finding: the prevailing paradigm, anchored in adaptations of Automatic Speech Recognition (ASR) and purely data-driven machine learning, is fundamentally inadequate for the task. As rigorously documented, these approaches are hamstrung by intrinsic misalignment—prioritizing word recognition over qualitative assessment—and besieged by practical impediments, including catastrophic demographic biases, a crippling dependency on unattainably perfect training data, and a failure to deliver the granular, diagnostic feedback essential for effective learning.

The critique of existing systems, however, illuminates the path forward. The very nature of the Quranic text—its immutable composition, precisely governed recitation rules (*Tajweed*), and meticulously preserved canonical readings (*Qira'at*)—presents a unique computational advantage. It invites a paradigm shift from statistical approximation to principled, knowledge-based analysis. We therefore conclude that the future of robust and equitable automated evaluation lies in architecting systems that are deeply informed by the Islamic sciences of *Tajweed*.

Ultimately, the goal of "Advancing Quranic Recitation through Technology" is not to replace the revered teacher-student tradition (*Talaqqi*) but to create digitally-augmented tools that extend its reach and accessibility. By grounding computational design in the timeless principles of *Tajweed*, technology can finally begin to serve this sacred tradition with the fidelity, precision, and respect it demands, empowering a new generation of learners worldwide. "'